\pdfoutput=1
\documentclass{article}

\usepackage{genai4health_2026}

\usepackage[utf8]{inputenc}
\usepackage[T1]{fontenc}
\usepackage{hyperref}
\hypersetup{hidelinks}
\usepackage{url}
\usepackage{booktabs}
\usepackage{amsfonts}
\usepackage{amsmath}
\usepackage{nicefrac}
\usepackage{microtype}
\usepackage[table]{xcolor}
\usepackage{graphicx}
\usepackage{multirow}
\usepackage{longtable}
\usepackage{placeins}

\input{GenAI4H/preamble/macros}

\title{GlucoAgent: Personalized Glucose Forecasting from Multimodal Personal Context}

\newcommand{\authorblock}[1]{\makebox[0.4\textwidth]{#1}}
\author{
    \authorblock{Xingzhi Sun\thanks{Work done at Meta.}}\\
    Yale University
    \And
    \authorblock{Hyo Jin Kim}\\
    Meta Reality Labs Research
    \AND
    \authorblock{Zhiyuan Wang}\\
    Meta Reality Labs Research
    \And
    \authorblock{Hao Chen}\\
    Meta Reality Labs Research
}

\newcommand{\best}[1]{\textbf{#1}}
\newcommand{\runnerup}[1]{\underline{#1}}

\begin{document}

\maketitle

\begin{abstract}
  Continuous glucose monitors, smartwatches, meal logs, and health profiles form a rich multimodal personal context that can significantly enhance glucose forecasting.
However, existing methods struggle to effectively utilize these signals to improve glucose prediction because the relevance of different modalities varies drastically across individuals and time.
We present \agentname{}, an LLM-based agentic system that addresses this challenge through dynamic, case-by-case analysis at inference time.
For each forecast, \agentname{} writes and runs code over an individual's multimodal context and returns predictions with a reason and supporting evidence.
The generated programs are highly diverse, ranging from retrieving similar past meals to fitting and back-testing physiological and statistical models.
On CGMacros, a multimodal glucose dataset, \agentname{} outperforms time series foundation models on every forecasting horizon, achieving an average RMSE of 22.5~mg/dL compared to 23.5~mg/dL of Chronos-2, the leading multimodal time series foundation model.
Furthermore, \agentname{} is highly effective at leveraging multimodal data. The addition of multimodal data reduces its average RMSE by 1.4~mg/dL (24.0~mg/dL to 22.5~mg/dL), compared to a marginal 0.3~mg/dL (23.8~mg/dL to 23.5~mg/dL) improvement for Chronos-2.
This advantage is most pronounced in challenging post-meal windows, where \agentname{} uses multimodal context to reduce 60-minute forecast RMSE by 4.6~mg/dL %
(31.0~mg/dL to 26.5~mg/dL), far exceeding the 0.8~mg/dL (31.9~mg/dL to 31.1~mg/dL) reduction achieved by Chronos-2. %

\end{abstract}

\section{Introduction}
\begin{figure}[t]
  \centering
  \includegraphics[width=\linewidth]{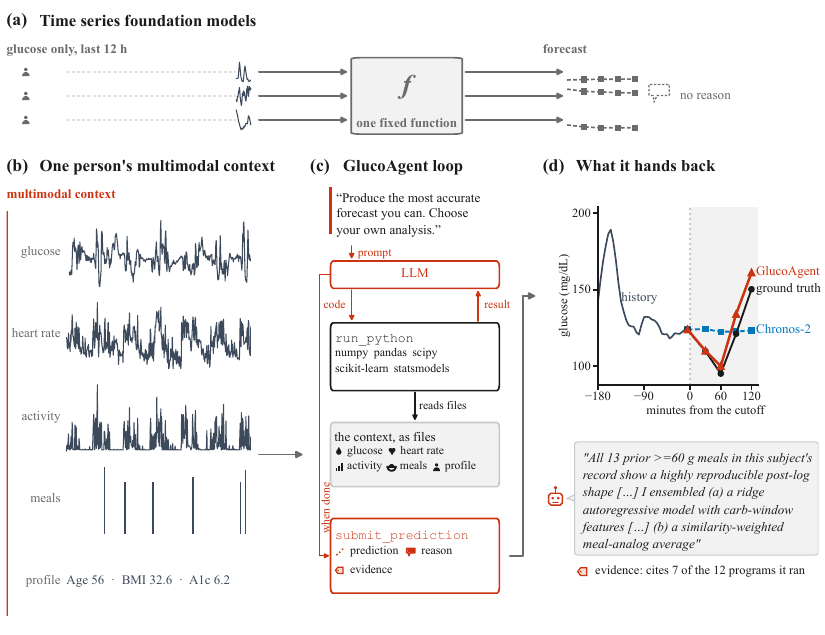}
  \caption{\textbf{One person’s multimodal context enters \agentname{}. The agent writes and runs code to construct a case-by-case analysis, then submits a forecast with a reason and supporting evidence.} \textbf{(a)}~Time series foundation models receive the last
  twelve hours of glucose and return a forecast without a
  reason. \textbf{(b)}~\Agentname{} receives one person's glucose, heart rate,
  activity, logged meals and health profile. \textbf{(c)}~The agent writes and
  runs code over that context, reads the outputs and continues
  until submission. \textbf{(d)}~It returns each forecast with a reason and
  supporting evidence. In this example, it combines responses to the
  person's past large meals with a regression fitted and validated on that
  person's history.}
  \label{fig:teaser}
\end{figure}

Glucose forecasting is becoming increasingly important as continuous glucose monitors (CGMs) provide dense and longitudinal glucose measurements for people with or without diabetes~\citep{zeevi2015,cgmacros}. 
Researchers have used CGM measurements to predict individual responses to meals and support personalized nutrition~\citep{zeevi2015}, while commercial insulin-pump systems use glucose forecasts to automatically adjust insulin delivery~\citep{battelino2017,forlenza2018,brown2019}. 
Therefore, accurate glucose forecasting is crucial for applications ranging from personalized metabolic-health management to automated diabetes care.

In this work, we study short-horizon glucose forecasting, which focuses on predicting continuous glucose monitor (CGM) measurements 30 to 120 minutes ahead.
An accurate forecast can provide advance warning of changes in glucose and act as a useful reference for users who need to control their glucose levels~\citep{woldaregay2019,zhu2021}.
Existing glucose forecasters range from physiological models of glucose--insulin dynamics to autoregressive models, neural networks and general-purpose time series foundation models~\citep{sparacino2007,woldaregay2019,zhu2021,chronos2,timesfm,moirai2,timer,glucofmbench}.
Forecasts have also been tailored to individuals by training a model for each participant~\mbox{\citep{attengluco,physiofeedbackagent}} or selecting a model based on validation error among different neural network architectures~\citep{personalizedllmglucose}.

Additional signals measured with smartwatches, meal logs, and health profiles can complement the CGM to provide a multimodal personal context. Because glucose is affected by an individual's physiology and daily activities, such multimodal personal context has the potential to significantly improve glucose prediction.
Multimodal forecasters have incorporated insulin and carbohydrates~\citep{metabonetbench}, wearable signals~\citep{attengluco,ssmcgm}, and health-record context~\citep{glullm}. 
Yet, existing methods have struggled to demonstrate substantial improvements after adding the multimodal context to glucose inputs.
For instance, in MetaboNet-Bench~\citep{metabonetbench}, adding insulin and carbohydrate inputs to five forecasting models reduced 30-minute RMSE by only 0.13 to 0.71~mg/dL. 
We argue that such limited improvements indicate underutilized multimodal personal context. This stems from a core challenge that the predictive utility of such context can vary across people and forecast moments.
For example, a person's earlier meal responses might be more helpful for forecasting near a meal, while forecasting near a quiet period may benefit more from recent glucose dynamics.
Across existing approaches, researchers determine the model class and the way multimodal context enters it during development. What remains underexplored is whether constructing the analysis case-by-case at forecast time can better address this challenge.

To study this question, we introduce \agentname{}, an LLM-based agentic system that constructs a tailored analysis from one person's multimodal context at the time of each forecast.
Building on language models that act through tools or executable programs~\citep{react,pal} and related agents that analyze personal health data~\citep{phia,wang2026pulse}, \agentname{} uses an agent loop that writes and runs code, reads the outputs, and continues until it submits a forecast with a reason and supporting evidence.
Because the agent can construct a different analysis for a different person or at a different time, this dynamic N-of-1 analysis provides flexible personalized analyses that use the multimodal personal context exactly as needed for each case.
The reason and supporting evidence returned with each forecast enable us to inspect the agent's logic, making our method more interpretable than a black-box forecasting model.
We equip the Python environment with scientific and statistical libraries and observe, through the generated reason and evidence, that the agent uses these libraries to build and test physiological models as well as statistical machine learning models depending on the scenario.
Figure~\ref{fig:teaser} illustrates how \agentname{} processes the multimodal personal context to make a forecast and submits a reason and supporting evidence.

On CGMacros~\citep{cgmacros}, a public multimodal glucose dataset, we compare \agentname{} with time series foundation models evaluated in GlucoFM-Bench~\citep{glucofmbench}.
With multimodal inputs, \agentname{} achieves a lower RMSE at every horizon from 30 to 120 minutes, with its widest lead of 1.6~mg/dL at 60 minutes compared with Chronos-2~\citep{chronos2}, the leading multimodal time series foundation model.
Across the four forecast horizons, adding the multimodal context reduces \agentnameposs{} RMSE by 1.1 to 2.1~mg/dL, compared with 0.1 to 0.7~mg/dL for Chronos-2.
To find out in which situations the multimodal context helps the most, we divided the prediction origins into post-meal, post-exercise, meal + exercise, and quiet situations.
The multimodal context, compared to using glucose only, helps \agentname{} most after meals, where it reduces the RMSE by 2.5~mg/dL on average, and by a massive 4.6~mg/dL at 60 minutes. In contrast, Chronos-2's RMSE is reduced by only 0.9~mg/dL on average and 0.8~mg/dL at 60 minutes.
On public glucose-only benchmarks, including the nine datasets in GlucoFM-Bench~\citep{glucofmbench}, \agentname{} remains competitive with time series foundation models and CGM-LSM, a foundation model trained specifically for glucose forecasting~\citep{cgmlsm}.

Our main contributions are as follows:
\begin{enumerate}
  \item We introduce \agentname{}, an LLM agentic system for personalized multimodal glucose forecasting that dynamically writes and runs code over a person's multimodal context for each forecast and submits a reason and supporting evidence.
  \item We demonstrate the efficacy of \agentname{} by comparing it against time series foundation models on the CGMacros dataset with multimodal records as well as the glucose-only datasets from GlucoFM-Bench~\citep{cgmacros,glucofmbench}.
  \agentname{} outperforms time series foundation models at every evaluated horizon with multimodal context, and matches their performance on glucose-only univariate settings.
  \item Through stratified evaluations across distinct physiological states (e.g., post-meal), qualitative case studies, and a systematic analysis of the generated code, we reveal when and how multimodal context improves predictions, offering transparency into the forecasting process.
\end{enumerate}

\section{Related Work}
\label{sec:related-work}

\paragraph{Personalized and multimodal glucose forecasting.}
Recent glucose forecasting methods include glucose-specific pretrained models~\citep{cgmlsm,gluformernature,glucofm} and general-purpose time series foundation models that perform zero-shot forecasting without glucose-specific training~\citep{chronos,chronos2,timesfm,moirai,moirai2,timer,timerxl}.
On GlucoFM-Bench~\citep{glucofmbench}, the best zero-shot time series foundation model, TimesFM-2.5~\citep{timesfm}, achieves forecast error within 5\% of the best fully supervised LSTM model of~\citet{martinsson2020} when both use glucose-only inputs.
Other methods tailor glucose forecasts to individuals by training a model for each participant~\citep{attengluco,physiofeedbackagent}, selecting a model for each participant based on validation error~\citep{personalizedllmglucose}, or updating a separate linear model as new CGM readings arrive~\citep{erenoruklu2009}.
Multimodal forecasters add heart rate and activity~\citep{attengluco,ssmcgm}, meal photographs~\citep{wolber2025}, or health-record context~\citep{glullm} to the glucose history.
MetaboNet-Bench~\citep{metabonetbench} also evaluates insulin and carbohydrate inputs, but finds that they reduce forecast error only slightly.
On CGMacros, \citet{physiofeedbackagent} forecast 120-minute after-meal glucose from multimodal inputs using a predictor fine-tuned for each participant.
A separate LLM agent uses these forecasts to propose meal changes.
Across personalized and multimodal glucose forecasting, researchers predetermine the model class and how multimodal context enters it during development, even when they fit or update a model separately for each participant.

\paragraph{Language models and agents for time series.}
Language models have been adapted to forecast time series through prompting or fine-tuning~\citep{llmtime,lstprompt,timellm,gpt4ts,calf}.
Some language-model forecasters also receive natural-language descriptions of the series, relevant events, or forecast constraints~\citep{contextiskey}, while retrieval-augmented forecasters use examples from other series~\citep{tsrag} or from the target series' own history~\citep{raft}.
General tool-using language-model methods include ReAct, which uses tool outputs to decide what to do next~\citep{react}, and PAL, which uses an interpreter to perform numerical calculations~\citep{pal}.
Time-series forecasting agents use language models to route among pretrained forecasters~\citep{moiraiagent} or coordinate prescribed roles, steps and model libraries~\citep{timeseriesscientist,castflow,dcats}.
For time-series analysis rather than forecasting, \citet{rechtorik} study coding agents that write and run Python code over time series to answer multiple-choice questions.

\paragraph{Agents for personal health.}
LLM agents have also been used to analyze personal health data.
Agents and other LLM-based methods for CGM retrieve relevant segments, summarize glucose patterns, compute retrospective metrics and answer questions~\citep{cgmagent,llmcgm,dmbench,healey2025}.
The Personal Health Agent includes a data-science agent that first produces an analysis plan and then writes and runs Python code over personal wearable and health-record data~\citep{personalhealthagent}.
PHIA writes and runs Python code over one person's wearable data to answer retrospective questions~\citep{phia}.
PULSE gives an agent purpose-built tools to choose which smartphone signals and look-back windows to inspect when predicting whether a person wants emotional support and is available to receive it~\citep{wang2026pulse}.
These methods analyze past or current health data.
Our \agentname{} writes and runs code over one person's multimodal context at each forecast origin to predict future glucose.

\section{Method}
\label{sec:method-agent}

\Agentname{} is an LLM-based agentic system that forecasts glucose from each person's multimodal context.
The overview of our system is illustrated in Figure ~\ref{fig:teaser}.
At each forecast origin, the agent writes and runs code over a person's context, then submits the forecasts with a reason and supporting evidence.

\subsection{Multimodal personal context}
Figure~\ref{fig:teaser}b summarizes the multimodal personal context supplied to \Agentname{} for each forecast.
Using CGMacros~\citep{cgmacros} as an example, this context encompasses multimodal time-series data (glucose, heart rate, energy expenditure, and metabolic equivalents), detailed meal logs (timestamps, meal types, and macronutrient contents), and a static health profile (demographics, body measurements, and laboratory results). While the dataset also contains meal photographs, we defer raw image analysis to future work.
We strictly retain only the information available prior to the forecast origin to enforce causal boundaries, and compile these records, alongside the target prediction timestamps, into a single \texttt{JSON} case file (see Appendix~\ref{app:agent-case-file} for formatting details). Notably, we inject no hand-engineered features. Instead, we provide the raw data, relying entirely on the agent to autonomously determine which contextual elements are relevant and how to process them for each individual case.
For glucose-only evaluations, the same agent receives a reduced case file containing only glucose and the target specification (Appendix~\ref{app:agent-glucose-only}).

\subsection{Per-origin numerical analysis}

We provide the agent a prompt that defines the forecasting task, available evidence, and a list of valid actions.
In the prompt, we also instruct it to ``Produce the most accurate forecast you can. Choose your own analysis.''
The prompt does not specify any forecasting model, feature set, or look-back window.
The complete prompt is provided in Appendix~\ref{app:agent-rendered-prompts}.

We equip the agent with two tools: a Python interpreter and a forecast-submission tool.
Following ReAct and program-aided reasoning~\citep{react,pal},
the agent writes a program, observes its output, and decides what to run next until it submits a forecast, as shown in Figure~\ref{fig:teaser}c.
Each forecast episode allows at most \maxturns{} LLM turns.
The Python tool is disabled in the final two turns, which are reserved for submission and correction if the first submission is invalid.
The Python programs run in a sandbox with read-only access to the data file, no network access, and fixed time and memory limits.
The Python environment provides general-purpose numerical and statistical packages, including NumPy, SciPy, pandas, scikit-learn, and statsmodels.
It also provides two optional Python modules for loading the case and reporting the available fields and their coverage.
We provide no pretrained glucose forecaster, time series foundation model, or modality-specific analysis pipeline.
Thus, the agent constructs its forecasting procedure using general-purpose computation; depending on the case, it may process the data, search the person's history for comparable events, fit models, and validate their predictions.

All reported experiments use \llmexact{} at high reasoning effort, fixed before test scoring and unchanged across benchmarks without dataset-specific prompts or tuning.
Appendix~\ref{app:agent-design} gives the complete runtime configuration, and Appendix~\ref{app:agent-leakage} describes the sandbox and temporal restrictions.

\subsection{Forecast submission with reasons and supporting evidence}

The forecast-submission tool requires one glucose value for each target timestamp, a reason for the forecast, and evidence supporting that reason.
Each successful Python call returns an evidence reference linked to the current case, the program, and its output.
An incomplete submission or invalid reference produces an error that the agent may correct and retry (Appendix~\ref{app:agent-evidence}).
Figure~\ref{fig:teaser}d shows \Agentname{}'s output for one forecast: predicted values together with a reason and cited evidence that make its underlying analysis inspectable.
In Section~\ref{sec:results-behaviour}, we use the submitted reason and evidence to reconstruct individual forecasts and examine how the agent uses multimodal personal context across people and forecast origins.

\section{Results}
\label{sec:results}

\subsection{Experimental setup}
\label{sec:results-setup}

We use CGMacros~\citep{cgmacros}, a public dataset in which each participant wore a continuous glucose monitor and a smartwatch while logging their meals.
We held out nine of the 45 participants as the test set, with three in each A1c range: normal, prediabetes, and diabetes. 
These participants provide 1,903 forecast origins spaced one hour apart.
At each origin, a method predicts glucose at 30, 60, 90, and 120 minutes in the future.
The multimodal context contains glucose, heart rate, activity, logged meals, and a baseline health profile.
Across the forecast origins, the available context has a median of 5.5 days of history and 22 logged meals (for participant and data details, see Appendix~\ref{sec:data-preparation-subsampling}).

We evaluate under three input settings: the last twelve hours of glucose, the full glucose history, and the multimodal personal context.
Comparing the first two settings shows the effect of longer glucose history, while comparing the second and third shows the additional effect of the non-glucose signals.
In the glucose-only settings, we evaluate \agentname{} and four time series foundation models, Chronos-2, TimesFM-2.5, Moirai-2.0, and Timer~\citep{chronos2,timesfm,moirai2,timer}. We also include persistence, a naive baseline with no parameters that repeats the last glucose reading as the forecast at every horizon.
Among these models, Chronos-2 accepts additional multimodal covariates and is therefore used for the multimodal comparison.
Chronos-2 receives heart rate, activity, meal nutrient values, and one-hot-encoded meal type as multivariate numerical time series, while \agentname{} receives the \texttt{JSON} context file described in Section~\ref{sec:method-agent} (for the exact inputs and baseline configurations, see Appendices~\ref{app:evaluation-guarantees} and~\ref{sec:tsfm-baseline-setup}).

Following~\citet{glucofmbench}, we compute root-mean-square error (RMSE) separately for each participant and horizon and then average across participants with equal weight.
Mean absolute error and further evaluation details are reported in Appendix~\ref{sec:additional-mae}.

\subsection{\Agentname{} outperforms Chronos-2 with the multimodal context}
\label{sec:results-cgmacros}
Table~\ref{tab:main-cgmacros-holdout} presents the evaluation results on the CGMacros dataset. 
With the multimodal context, our \agentname{} achieves lower RMSE than Chronos-2~\citep{chronos2} at all four horizons.
Across the four horizons, its mean RMSE is 22.5~mg/dL, compared with 23.5~mg/dL for Chronos-2.
The lead is largest at 60 minutes, where the two methods have RMSEs of 22.1 and 23.7~mg/dL.

\begin{figure}[t]
  \centering
  \includegraphics[width=\linewidth]{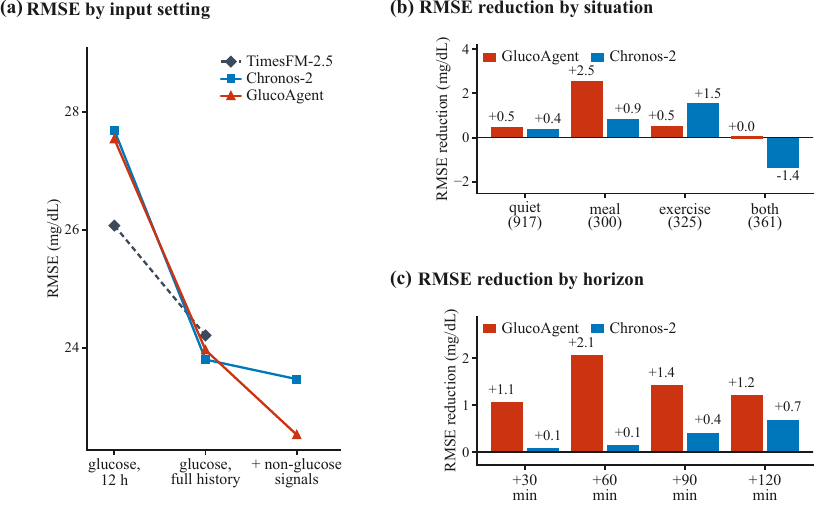}
  \caption{\textbf{The multimodal context reduces \agentnameposs{} RMSE more than Chronos-2's.}
  \textbf{(a)}~Mean participant-level RMSE over the four horizons as the input expands from twelve hours of glucose to the full glucose history and then the multimodal context.
  \textbf{(b)}~The RMSE reduction from the multimodal context in four forecast situations.
  \textbf{(c)}~The same reduction at each forecast horizon.}
  \label{fig:results}
\end{figure}

For both methods, incorporating a longer glucose history improves performance.
Expanding the context window from twelve hours to the full glucose history reduces the four-horizon mean RMSE by 3.6~mg/dL for \agentname{} and 3.9~mg/dL for Chronos-2.
When using only this full glucose history, \agentname{} and Chronos-2 perform competitively, differing by a maximum of 0.3~mg/dL at any given horizon. However, incorporating the multimodal context reduces \agentnameposs{} four-horizon mean RMSE by 1.4~mg/dL, compared with 0.3~mg/dL for Chronos-2 (Figure~\ref{fig:results}a).

\begin{table}[t]
  \caption{\textbf{Adding the multimodal context reduces \agentnameposs{} RMSE more than Chronos-2's, and \agentname{} has lower RMSE with the multimodal context at all four horizons.} Participant-level RMSE in mg/dL, mean (standard deviation) over the nine participants. \textbf{Bold} marks the lowest and \underline{underlining} the second lowest error within a setting ($^{\dagger}$Timer's input handling is described in Appendix~\ref{sec:tsfm-baseline-setup}).}
  \label{tab:main-cgmacros-holdout}
  \centering
  \setlength{\tabcolsep}{4pt}
  \begin{tabular*}{\textwidth}{@{\extracolsep{\fill}}lrrrr}
    \toprule
    Method & 30 min & 60 min & 90 min & 120 min \\
    \midrule
    \multicolumn{5}{l}{\textit{Reference}} \\
    \quad Persistence & 20.429 (5.574) & 29.151 (11.147) & 34.646 (14.176) & 38.003 (16.954) \\
    \addlinespace
    \multicolumn{5}{l}{\textit{Glucose only, 12 hours of history}} \\
    \quad TimesFM-2.5 & \best{17.853} (5.027) & \best{25.586} (10.228) & \best{29.495} (12.924) & \best{31.354} (15.047) \\
    \quad Moirai-2.0 & 18.446 (5.625) & \runnerup{27.011} (10.841) & 31.587 (14.169) & 33.886 (16.853) \\
    \quad Timer$^{\dagger}$ & 20.089 (6.603) & 27.342 (11.607) & \runnerup{30.779} (14.495) & \runnerup{31.559} (16.221) \\
    \quad Chronos-2 & \runnerup{18.446} (5.070) & 27.137 (10.362) & 31.449 (13.229) & 33.720 (16.117) \\
    \quad \agentname{} & 18.679 (5.080) & 27.209 (10.093) & 31.077 (12.579) & 33.191 (14.883) \\
    \addlinespace
    \multicolumn{5}{l}{\textit{Glucose only, full history}} \\
    \quad TimesFM-2.5 & \best{17.278} (4.740) & \runnerup{24.106} (9.513) & 27.110 (12.378) & 28.346 (14.336) \\
    \quad Moirai-2.0 & 17.437 (5.061) & 24.353 (9.545) & 27.467 (11.972) & 28.837 (13.843) \\
    \quad Timer$^{\dagger}$ & 18.057 (5.528) & 24.672 (10.105) & 27.539 (12.431) & 28.757 (14.145) \\
    \quad Chronos-2 & \runnerup{17.329} (4.906) & \best{23.853} (9.328) & \best{26.464} (11.342) & \best{27.553} (13.031) \\
    \quad \agentname{} & 17.416 (4.930) & 24.187 (9.150) & \runnerup{26.584} (10.846) & \runnerup{27.661} (12.395) \\
    \addlinespace
    \multicolumn{5}{l}{\textit{Multimodal personal context}} \\
    \quad Chronos-2 & 17.246 (5.125) & 23.711 (9.586) & 26.052 (11.527) & 26.866 (12.887) \\
    \quad \agentname{} & \best{16.357} (3.605) & \best{22.132} (8.220) & \best{25.163} (9.663) & \best{26.450} (11.333) \\
    \bottomrule
  \end{tabular*}
\end{table}

\subsection{The improvement from the multimodal context is largest after meals}
\label{sec:results-situations}

We investigate whether the benefit of the multimodal context varies with the event preceding the forecast origin.
For this retrospective analysis, we classify each forecast origin into one of four situations based on the meal and exercise information available before it:
quiet, post-meal, post-exercise, or meal + exercise
(for the definitions and number of origins in each situation, see Appendix~\ref{sec:data-preparation-subsampling} and Figure~\ref{fig:by-horizon}a).

The full evaluation results for each situation and horizon are reported in Figure~\ref{fig:by-horizon} and Tables~\ref{tab:cgmacros-holdout-stratum-quiet}--\ref{tab:cgmacros-holdout-stratum-compound}.

The largest improvement for \agentname{} occurs after meals.
Adding the multimodal context reduces its RMSE by 2.5~mg/dL in the post-meal situation, compared with 0.5~mg/dL when quiet, 0.5~mg/dL after exercise, and 0.0~mg/dL when a meal and exercise are both present (Figure~\ref{fig:results}b).
Chronos-2 improves most after exercise, by 1.5~mg/dL, and its post-meal improvement is 0.9~mg/dL.

The performance gains from the multimodal context also vary by forecast horizon.
For \agentname{}, the largest improvement happens at 60 minutes, where the multimodal context reduces RMSE by 2.1~mg/dL, compared with 0.1~mg/dL for Chronos-2 (Figure~\ref{fig:results}c).
Within the post-meal situation, the reduction for \agentname{} reaches 4.6~mg/dL at 60 minutes.

The large improvement in the post-meal situation motivates us to examine how the agent used meal information and the rest of the personal history in individual forecasts.

\subsection{\Agentname{} uses different analyses across forecast cases}
\label{sec:results-behaviour}

\begin{figure}[t]
  \centering
  \includegraphics[width=\linewidth]{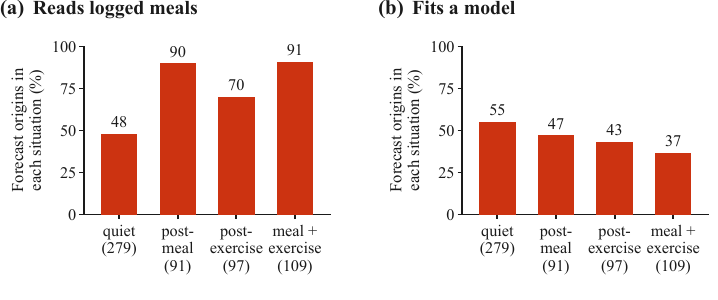}
  \caption{\textbf{The cited programs differ across forecast situations.}
  Each bar shows the percentage of origins in a situation for which the cited programs read the logged meals or fit a model.
  The subsample contains 576 origins: 279 quiet, 91 post-meal, 97 post-exercise, and 109 meal + exercise.}
  \label{fig:behavior-main}
\end{figure}

The programs that \agentname{} writes and runs differ across forecast situations.
To understand the behavior, we examined a 576-origin subsample stratified by participant and forecast situation to reflect the full test set (Appendix~\ref{sec:data-preparation-subsampling}).
In this subset, \agentname{} reads the meal log in 90\% of post-meal cases and 48\% of quiet cases.
As shown in Figure~\ref{fig:behavior-main}, \agentname{} fits a model in 55\% of quiet cases and 37\% of cases in the meal + exercise situation.
Meal-log use is most common after meals, whereas model fitting is most common during quiet periods.

\begin{figure}[p]
  \centering
  \includegraphics[width=\linewidth]{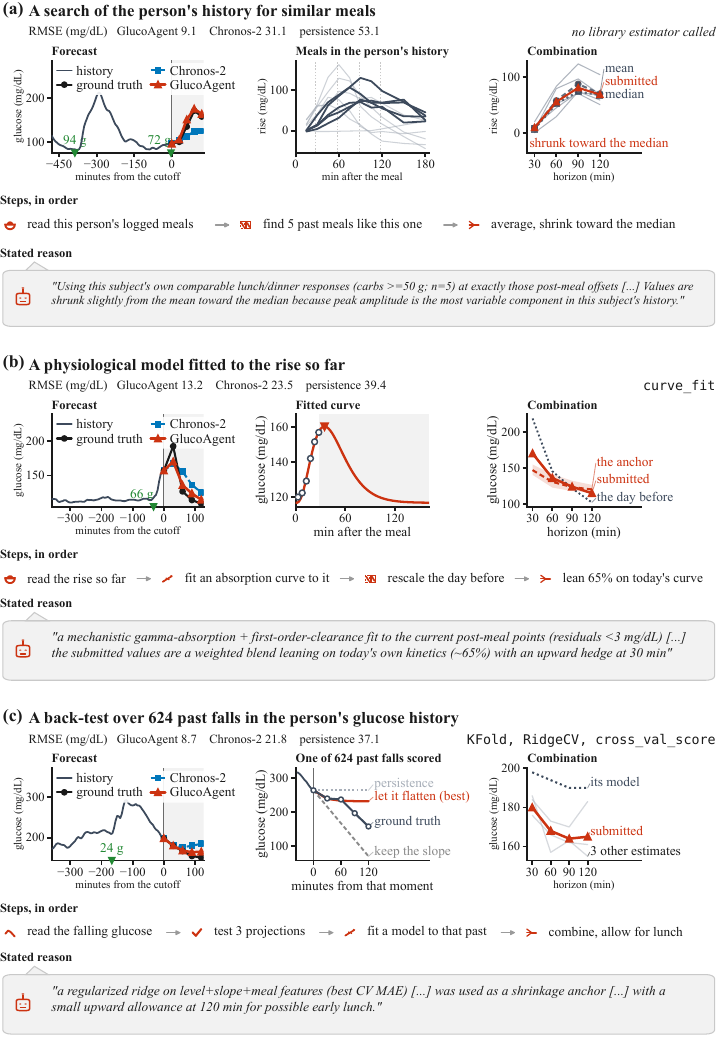}
  \caption{\textbf{Three case-by-case analyses constructed from personal history.}
  Each row reconstructs one forecast from the agent's cited programs and submitted reason.
  The rows show a search for similar past meals, a physiological model fitted to an ongoing rise, and three rules for continuing a falling glucose trend back-tested on earlier windows.}
  \label{fig:cases-main}
\end{figure}

Next, we examine individual cases.
A procedure recorded before case inspection identified nine accurate forecasts with substantial glucose changes.
Figure~\ref{fig:cases-main} reconstructs three from the submitted reason and cited programs, while Figures~\ref{fig:cases-appendix-1} and~\ref{fig:cases-appendix-2} show the other six.

For the forecast in Figure~\ref{fig:cases-main}a, a meal is logged three minutes before the forecast origin, before its glucose response can be observed.
The agent reads all twelve meals in the participant's meal history, keeps the five lunches and dinners with more than 50~g of carbohydrate, and uses their mean and median responses as anchors for its forecast.

In Figure~\ref{fig:cases-main}b, the agent fits a physiological model of absorption and clearance to the first six measurements of an ongoing post-meal rise.
It compares the fitted curve with the same breakfast on the previous day.
The agent gives more weight to the current curve because the rise is already slowing and the current breakfast contains more protein, and it discounts the earlier day's post-peak decline because the decline coincided with exercise.

In Figure~\ref{fig:cases-main}c, the agent examines a quiet period with a steep glucose decline.
It finds 624 earlier windows that meet its falling-slope threshold and back-tests three rules: hold the current value, continue the current slope, and let the slope decay exponentially.
The exponentially damped rule has the lowest mean absolute error in these back-tests.
The final forecast combines this rule with nearest-neighbor, historical-analogue, and ridge-regression estimates.

Together, these case studies show how \agentname{} personalizes each forecast. The agent decides which parts of the person's multimodal context to use, how to form  predictions from them, and how to validate and combine those predictions into the final forecast.
More broadly, across all 1{,}903 held-out origins, the agent runs 14{,}468 successful programs, with a median of seven per origin, and writes its own function or class in 64\% of the origins.
In the 576-origin subsample, the signals, estimators, validation checks, and presence of an agent-written function form 186 different combinations.
Of these, 120 occur only once (for the full analysis of the programs, see Appendix~\ref{sec:behavior-appendix}).

\subsection{Accuracy depends on the language model and reasoning effort}
\label{sec:results-configuration}

We observed that the performance varies substantially depending on the language model and the reasoning effort.
We vary the language model and reasoning effort while keeping the multimodal input setting and the rest of the method fixed (Table~\ref{tab:llm-multimodal-ablation}).
The \llmexact{} configuration at high reasoning effort has the lowest RMSE at every horizon.
The alternatives increase the four-horizon mean RMSE by 1.1 to 5.2~mg/dL.
The RMSE reduction from the multimodal context also depends on the model: it is 1.4~mg/dL for \llmexact{} at high reasoning effort and 0.3~mg/dL for Claude Sonnet 5 at high reasoning effort (for the complete input comparison, see Appendix~\ref{sec:additional-llm-ablation}).

\begin{table}[htbp]
  \caption{\textbf{\llmexact{} at high reasoning effort has the lowest RMSE at every horizon among the tested configurations.} Participant-level RMSE in mg/dL, mean (standard deviation) over the nine participants. \textbf{Bold} marks the lowest and \underline{underlining} the second lowest error at each horizon.}
  \label{tab:llm-multimodal-ablation}
  \centering
  \setlength{\tabcolsep}{3.2pt}
  \resizebox{1.0\textwidth}{!}{%
  \begin{tabular}{llrrrr}
    \toprule
    Model & Effort & 30 min & 60 min & 90 min & 120 min \\
    \midrule
    Claude Opus 5 (ours) & high & \textbf{16.357} (3.605) & \textbf{22.132} (8.220) & \textbf{25.163} (9.663) & \textbf{26.450} (11.333) \\
    Claude Opus 5 & medium & \underline{17.016} (3.683) & 23.746 (8.594) & 27.308 (10.659) & 28.999 (12.550) \\
    Claude Sonnet 5 & high & 18.595 (4.177) & 26.613 (8.857) & 31.108 (11.528) & 34.511 (14.200) \\
    Gemini 3.1 Pro & high & 18.964 (5.404) & 25.389 (10.162) & 29.277 (11.128) & 30.693 (13.674) \\
    GPT-5.6-sol & high & 17.202 (4.263) & \underline{23.300} (8.562) & \underline{26.436} (9.079) & \underline{27.753} (10.995) \\
    \bottomrule
  \end{tabular}}
\end{table}

\subsection{\Agentname{} remains competitive on glucose-only benchmarks}
\label{sec:results-benchmarks}

Our main task uses the multimodal personal context, while the glucose-only benchmarks below test whether \agentname{} remains competitive when given only glucose history.
Under the GlucoFM-Bench protocol~\citep{glucofmbench}, each method receives twelve hours of glucose history and predicts 30 minutes ahead on nine public CGM datasets.
We use a fixed subset of at most 32 origins per participant for a total of 8{,}646 origins.
The relative performance of the time series foundation models is similar on the subset and the full population (Table~\ref{tab:appendix-glucofm-subset-full-tsfm}).

\Agentname{} obtains the lowest RMSE on five of the nine datasets and the second lowest on two more (Table~\ref{tab:main-glucofm-30min}).
Where another method has lower error, the difference is below 0.8~mg/dL.
Our glucose-only baseline runs are compared with their published results in Appendix~\ref{sec:reproducing-published-results}.
On OhioT1DM~\citep{marling2020} under the CGM-LSM protocol, \agentname{} is within 0.35~mg/dL of the best time series foundation model at the 30-, 60-, and 120-minute horizons (Table~\ref{tab:appendix-cgmlsm} in Appendix~\ref{sec:additional-cgmlsm}).
The table also includes the published CGM-LSM result, from a foundation model trained specifically for glucose forecasting~\citep{cgmlsm}.
Although that result uses a different evaluation population, \agentname{} has lower reported error at 120 minutes.

\begin{table}[t]
  \caption{\textbf{With its configuration unchanged, \agentname{} has the lowest RMSE on five of nine datasets and the second lowest on two.} Forecasting under the GlucoFM-Bench protocol: twelve hours of glucose history, a 30-minute horizon and participant-level RMSE in mg/dL as mean (standard deviation), on at most 32 origins per participant. \textbf{Bold} marks the best and \underline{underlining} the second best method on each dataset.}
  \label{tab:main-glucofm-30min}
  \centering
  \setlength{\tabcolsep}{2.4pt}
  \resizebox{1.0\textwidth}{!}{%
  \begin{tabular}{lrrrrrr}
    \toprule
    Dataset & Persistence & Chronos-2 & Moirai-2.0 & Timer & TimesFM-2.5 & \agentname{} \\
    \midrule
    Hall2018
      & 13.90 (7.75) & 13.27 (6.72) & \runnerup{12.96} (6.94) & 14.27 (8.10) & \best{12.92} (6.49) & 13.59 (6.21) \\
    BIG IDEAS Lab
      & 15.63 (4.04) & \runnerup{14.21} (3.84) & 14.29 (3.97) & 15.95 (5.19) & \best{13.78} (3.80) & 14.51 (3.71) \\
    ShanghaiT1DM
      & 17.64 (8.13) & 12.51 (5.27) & \runnerup{11.89} (4.56) & 19.59 (9.00) & 12.06 (5.22) & \best{11.04} (4.75) \\
    ShanghaiT2DM
      & 15.03 (5.02) & 11.77 (3.80) & 11.47 (3.51) & 15.40 (5.76) & \runnerup{11.14} (3.66) & \best{11.13} (3.57) \\
    HUPA-UCM
      & 21.58 (8.19) & 18.04 (7.57) & \runnerup{17.12} (6.86) & 24.49 (7.81) & 17.35 (7.42) & \best{16.35} (6.69) \\
    T1DM-UOM
      & 25.46 (6.11) & \runnerup{22.01} (5.53) & 22.62 (6.08) & 24.99 (6.47) & 22.34 (5.83) & \best{21.24} (5.45) \\
    Bris-T1D Open
      & 20.87 (9.82) & 18.31 (8.33) & 18.05 (7.94) & 21.35 (10.26) & \runnerup{17.89} (8.44) & \best{17.58} (8.61) \\
    AZT1D
      & 22.34 (5.74) & 20.52 (5.61) & 20.30 (5.61) & 23.17 (5.42) & \best{19.69} (5.15) & \runnerup{20.07} (5.98) \\
    OhioT1DM
      & 24.16 (4.53) & 20.77 (4.90) & \best{20.28} (4.65) & 25.46 (5.41) & 20.88 (4.50) & \runnerup{20.76} (5.99) \\
    \bottomrule
  \end{tabular}}
\end{table}

\FloatBarrier
\section{Conclusion}
\label{sec:conclusion}

In this work, we introduced \agentname{}, an LLM-based agentic system for personalized multimodal glucose forecasting. Traditional models struggle to integrate multimodal context because its predictive utility varies drastically across individuals and time. \agentname{} overcomes this limitation by replacing static prediction functions with dynamic, case-by-case computational analysis. At each prediction point, the agent writes, executes, and iterates on bespoke Python code to analyze an individual's specific health and behavioral context.
Empirically, \agentname{} outperforms the leading time-series foundation model, Chronos-2, across all forecast horizons on the CGMacros dataset. In particular, \agentname{} demonstrates a superior ability to extract signal from multimodal records beyond just the glucose data, resulting in large error reductions in challenging post-meal windows where baseline models fail. Our analysis of the agent’s generated code reveals sophisticated, autonomous investigation strategies ranging from historical meal retrieval to physiological model fitting and rigorous back-testing. While our current multimodal evaluation focuses on CGMacros due to the scarcity of comprehensive, open-source datasets, future work will scale this agentic approach to broader, more diverse populations as new multimodal cohorts emerge.

\bibliographystyle{plainnat}
\bibliography{GenAI4H/references}

\clearpage
\appendix

\section{Agent Design in Detail}
\label{app:agent-design}

This appendix records the exact configuration of \agentname{}: the runtime
settings, the interface, the instruction text, the action schemas, the
sandbox, the control loop, the validation rules, and the contents of the
case file.  Everything here describes one configuration that was
hash-pinned before any held-out origin was scored and not changed
afterwards.  What the agent \emph{did} with this interface is reported in
the results.

\subsection{Frozen configuration}
\label{app:agent-runtime}

The forecasting policy is a large language model with a Python environment
and a structured submission action.
Table~\ref{tab:agent-runtime-settings} records the frozen inference and
execution settings, and Table~\ref{tab:agent-isa} states the interface as
inputs, state, actions and termination.

\begin{table}[htbp]
  \centering
  \small
  \begin{tabular}{p{0.46\linewidth}p{0.46\linewidth}}
    \toprule
    Component & Frozen setting \\
    \midrule
    Language model & \llmexact{} \\
    Provider model identifier & \llmexactid{} \\
    Transport & Bedrock Converse-compatible proxy \\
    Reasoning setting & High, adaptive \\
    Sampling temperature and top-$p$ & Not set; provider defaults \\
    Model calls per origin & at most \maxturns{} \\
    Python programs per origin & at most \maxcodecalls{} \\
    Generated tokens per model call & at most \maxouttokens{} \\
    Python & 3.9.25 \\
    NumPy / SciPy / pandas & 1.26.4 / 1.13.0 / 2.2.2 \\
    scikit-learn / statsmodels & 1.4.2 / 0.14.2 \\
    Network access & Disabled \\
    Wall-clock and CPU limit per program & \sandboxtimeout{} s each \\
    Address space per program & \sandboxmem{} MiB \\
    Submitted source per program & at most \maxcodebytes{} bytes \\
    Captured bytes per output stream & \maxcapturebytes{} \\
    \bottomrule
  \end{tabular}
  \caption{Frozen runtime settings for the multimodal forecasting agent.}
  \label{tab:agent-runtime-settings}
\end{table}

\begin{table}[htbp]
  \centering
  \small
  \begin{tabular}{p{0.16\linewidth}p{0.76\linewidth}}
    \toprule
    Element & Definition \\
    \midrule
    Inputs & The supplied context up to the cutoff, covering glucose, heart
      rate, activity, logged meals and the
      baseline profile, together with the timestamps of the meal
      photographs. \\
    State & The accumulated messages and a case-local filesystem.  Files
      persist across programs; each program starts a fresh interpreter. \\
    Actions & Run Python code, or submit one glucose value per
      requested timestamp with a reason and, after any
      successful program, supporting evidence references. \\
    Termination & The first valid submission is accepted; otherwise the
      episode ends when the model-call budget is exhausted. \\
    \bottomrule
  \end{tabular}
  \caption{Inputs, state, actions and termination condition.}
  \label{tab:agent-isa}
\end{table}

\subsection{Agent-visible data}
\label{app:agent-case-file}

The case file is mounted read-only at \texttt{/input/case.json}.  Its outer
object carries the requested horizons and target timestamps; its
\texttt{case} object carries an opaque case identifier, the participant
identifier as recorded in the source study, the exclusive cutoff, the start
of the supplied history, a columnar time series, an event list, the timestamps
of the meal photographs, the baseline profile, and provenance
digests of the source archive.

The nominal forecast origin is the start of the last visible five-minute
interval, and the exclusive end of that interval is the information cutoff.
The four targets are 30, 60, 90 and 120 minutes after the origin, and
therefore 25, 55, 85 and 115 minutes after the cutoff.  The exact
timestamps are supplied to the agent, so the labeling convention creates
no mismatch between what is predicted and what is scored, and it is applied
identically to every method compared.

The time series is columnar: parallel arrays of interval start and
exclusive end, plus one array per signal.  The eight signals are
\texttt{glucose\_mg\_dl}, \texttt{heart\_rate\_bpm},
\texttt{activity\_energy\_kcal}, \texttt{mets}, \texttt{meal\_carbs\_g},
\texttt{meal\_fat\_g}, \texttt{meal\_fiber\_g} and
\texttt{meal\_protein\_g}.  Meal values are sparse on the grid; the
separate event list retains the event timestamp, the meal type and the same
macronutrients.  Missing values are JSON nulls.  All history ending at the
cutoff is included, and the method imposes no look-back window.
Table~\ref{tab:agent-case-contents} summarizes what a case file holds.

\begin{table}[htbp]
  \centering
  \small
  \begin{tabular}{lrrr}
    \toprule
    Quantity & Minimum & Median & Maximum \\
    \midrule
    Pre-cutoff history (hours)      & 25.5 & 132.0 & 237.1 \\
    Five-minute rows                & 306  & 1{,}584 & 2{,}845 \\
    Logged meal events              & 3    & 22    & 61 \\
    Glucose coverage                & 0.92 & 0.99  & 1.00 \\
    Heart-rate coverage             & 0.56 & 0.99  & 1.00 \\
    Activity-energy coverage        & 0.90 & 1.00  & 1.00 \\
    METs coverage                   & 0.00 & 1.00  & 1.00 \\
    \bottomrule
  \end{tabular}
  \caption{Contents of the case file across the 576-origin subsample of
  held-out CGMacros forecasts.  Coverage is the fraction of five-minute rows carrying
  an observation for that signal.  METs are absent entirely for some
  participants, which is why the minimum is zero.}
  \label{tab:agent-case-contents}
\end{table}

The photograph entries tell the agent only that a meal photograph was taken
and when; they are subject to the same cutoff check as every other timestamped input
(Section~\ref{app:agent-leakage}), and no image bytes are mounted in the
workspace or attached to a model request.

\subsection{Vocabulary of the quoted prompt}
\label{app:agent-vocabulary}

The instruction below was written for the evaluation harness and uses its
internal vocabulary.  We quote it unchanged, because it is part of
the frozen configuration, and translate it here.  Read \emph{capability} as
helper module; \emph{workbench} and \emph{workbench index} as the set of
helper modules and the catalog listing them; \emph{packet} as the case
file; \emph{terminal} as the sandboxed workspace; \emph{toolbox} as the
Python package name under which the helper modules are importable;
\emph{candidate-owned} as supplied by us rather than by the model;
\emph{evidence\_ref} as an evidence reference; and \emph{case} as one
forecast origin.  Where the instruction promises that the workspace
contains no \emph{host repository}, \emph{run files}, \emph{truth} or
\emph{other cases}, it means, respectively, our source code, the harness's
own output files, the ground-truth glucose values, and every forecast
origin other than this one.

\subsection{Exact rendered prompts}
\label{app:agent-rendered-prompts}

A first model request contains a system instruction and a per-origin user
message.  The function schemas are supplied alongside those messages rather
than interpolated into them; Section~\ref{app:agent-tools} reproduces the two
model-facing action schemas separately.  We give complete rendered messages
for both configurations below.

\subsubsection{Multimodal system instruction}
\label{app:agent-system-prompt}

The following is the complete rendered system instruction, 2{,}232
characters long.  The block delimited by
\verb|<workbench_capabilities>| is the helper catalog of
Section~\ref{app:agent-helpers}, generated mechanically from the helper
modules rather than written by hand.

\begin{Verbatim}[breaklines=true,breakanywhere=true,fontsize=\scriptsize]
You are an autonomous glucose-forecasting agent. You have sandboxed scientific
Python through run_python. Every call starts a fresh Python 3.9 interpreter in the current case's
private /work directory; files under /work persist across calls, while interpreter variables and
processes do not. NumPy, SciPy, pandas, scikit-learn, and statsmodels are installed; no activation
or package installation is needed. Normal imports work, and pathlib may inspect or create files in
the private workspace when useful.

The cutoff-safe case is read-only at /input/case.json.

Optional candidate-owned workbench capabilities are listed below.
<workbench_capabilities>
- case_data
  Use when: loading the packet, aligned time series, or literal static profile.
  Import: from toolbox.case_data import load_packet, series_frame, static_pairs
  Entrypoints: load_packet(), series_frame(), static_pairs()
  Signatures:
    load_packet(path=CASE_PATH)
    series_frame(value=None)
    static_pairs(value=None, *, include_derived=True)
  Details: /work/toolbox/case_data.md
- modalities
  Use when: checking availability, coverage, missingness, or recency.
  Import: from toolbox.modalities import inventory
  Entrypoints: inventory()
  Signatures:
    inventory(value=None)
  Details: /work/toolbox/modalities.md
</workbench_capabilities>
Inspect or import only what is relevant. These capabilities do not prescribe a
forecasting method. You may create any safe files or folders under /work. The terminal has no network, credentials,
host repository, run files, truth, or other cases.

Produce the most accurate forecast you can. Choose your own analysis. Use only supplied case data and general methodological knowledge; never
invent unavailable episode facts. Prefer concise programs and print only results useful for the
next decision. Up to fourteen Python calls are available, but submit earlier when the
forecast is adequately supported. Every successful Python result includes an evidence_ref bound
to the exact code and output. When ready, call submit_prediction exactly once and cite the refs
that directly support the public reason in evidence_refs. Do not expose private chain-of-thought;
the final reason should be a short public summary.
\end{Verbatim}

Three properties of this text matter for interpreting the results.  It
names no forecasting method, no feature, no window length and no
physiological rule.  It states an objective and then delegates the analysis
in one sentence.  And it presents the helper modules as optional, giving a
contract and a condition of relevance but no recommendation to use them.

One instruction is stricter than the mechanism it describes: the text says
to ``call submit\_prediction exactly once'', whereas the harness allows a
rejected submission to be repaired and resubmitted
(Section~\ref{app:agent-evidence}).

\subsubsection{Multimodal per-origin task message}
\label{app:agent-user-message}

The instruction is followed by one short user message per origin.  One
rendered instance is reproduced below to show the exact format; only the
case identifier, the cutoff and the four target timestamps differ between
origins.

\begin{Verbatim}[breaklines=true,breakanywhere=true,fontsize=\scriptsize]
Forecast this case at the four scored horizons.

case_id=009_000101; cutoff_exclusive_s=1600621200
horizons_minutes=[30,60,90,120]
target_timestamps_s=[1600622700,1600624500,1600626300,1600628100]

/input/case.json is a JSON object with keys schema, horizons_minutes,
target_timestamps_s, and case. case.time_series is columnar with start_s,
end_exclusive_s, and fields; case.events contains cutoff-safe events; and
case.static_profile contains the supplied untimed baseline profile. Missing values are null.

The workbench index in the system context includes a copyable `Import:` line for each capability. When a listed `Use when:` condition matches analysis you intend to perform, prefer that `Import:` line over reimplementing the capability; ignore entries that do not match.

Use run_python whenever computation, Python file work, or inspection would help. Then submit values in
the same order as horizons_minutes: 30, 60, 90, 120.
\end{Verbatim}

The routing paragraph is the only guidance about helper use anywhere in the
instruction, and it is conditional: it asks the agent to prefer a listed
import over reimplementing the same functionality \emph{when the listed
condition matches an analysis it already intends to perform}, and to ignore
entries that do not match.  It does not say which analysis to perform.

\subsubsection{Glucose-only system instruction}
\label{app:agent-cgm-system-prompt}

The glucose-only configuration has a separately frozen rendered instruction.
It retains the same task sentence and execution rules, but its mechanically
rendered helper catalog exposes only the glucose series.  The complete
instruction is 2{,}125 characters long.

\begin{Verbatim}[breaklines=true,breakanywhere=true,fontsize=\scriptsize]
You are an autonomous glucose-forecasting agent. You have sandboxed scientific
Python through run_python. Every call starts a fresh Python 3.9 interpreter in the current case's
private /work directory; files under /work persist across calls, while interpreter variables and
processes do not. NumPy, SciPy, pandas, scikit-learn, and statsmodels are installed; no activation
or package installation is needed. Normal imports work, and pathlib may inspect or create files in
the private workspace when useful.

The cutoff-safe case is read-only at /input/case.json.

Optional candidate-owned workbench capabilities are listed below.
<workbench_capabilities>
- case_data
  Use when: loading the CGM-only packet or aligned glucose series.
  Import: from toolbox.case_data import load_packet, series_frame
  Entrypoints: load_packet(), series_frame()
  Signatures:
    load_packet(path=CASE_PATH)
    series_frame(value=None)
  Details: /work/toolbox/case_data.md
- modalities
  Use when: checking CGM coverage, missingness, or recency.
  Import: from toolbox.modalities import inventory
  Entrypoints: inventory()
  Signatures:
    inventory(value=None)
  Details: /work/toolbox/modalities.md
</workbench_capabilities>
Inspect or import only what is relevant. These capabilities do not prescribe a
forecasting method. You may create any safe files or folders under /work. The terminal has no network, credentials,
host repository, run files, truth, or other cases.

Produce the most accurate forecast you can. Choose your own analysis. Use only supplied case data and general methodological knowledge; never
invent unavailable episode facts. Prefer concise programs and print only results useful for the
next decision. Up to fourteen Python calls are available, but submit earlier when the
forecast is adequately supported. Every successful Python result includes an evidence_ref bound
to the exact code and output. When ready, call submit_prediction exactly once and cite the refs
that directly support the public reason in evidence_refs. Do not expose private chain-of-thought;
the final reason should be a short public summary.
\end{Verbatim}

\subsubsection{Glucose-only per-origin task message}
\label{app:agent-cgm-user-message}

This is a complete rendered CGMacros instance from the full-history
glucose-only evaluation.  The same renderer also supports the longer target
arrays used by the other glucose-only benchmarks.

\begin{Verbatim}[breaklines=true,breakanywhere=true,fontsize=\scriptsize]
Forecast this CGM-only case at every requested target.

case_id=cgmonly-7fcadbd2d2b3447a845cc615_000000; cutoff_exclusive_s=1588438800
forecast_offsets_minutes=[30,60,90,120]
target_timestamps_s=[1588440300,1588442100,1588443900,1588445700]
scored_horizons_minutes=[30,60,90,120]

/input/case.json contains only timestamps, glucose values, cutoff metadata, and target metadata.
Use run_python whenever computation or inspection would help. Submit values in exactly the same
order as target_timestamps_s.
\end{Verbatim}

The 12-hour, 144-sample glucose-only setting does not have a different prompt.
It uses the byte-identical system instruction and the same per-origin message
template.  For the matching origin above, only the opaque case identifier in
the rendered message changes; the cutoff, target arrays and all surrounding
text are identical.  The substantive experimental difference is in
\texttt{/input/case.json}: its timestamp and glucose arrays contain the final
144 samples rather than the available full history.

\subsection{Model-facing action schemas}
\label{app:agent-tools}

Exactly two actions are ever offered, with the schemas below;
\texttt{run\_python} is withheld in the final two model calls, as described
in Section~\ref{app:agent-loop}.  Across origins only the constant target
timestamps change, and the instance shown is the one belonging to the
message in Section~\ref{app:agent-user-message}.  The optional Python
imports advertised in the instruction are data-access helpers, not
function-calling actions.

\begin{Verbatim}[breaklines=true,breakanywhere=true,fontsize=\scriptsize]
{
  "type": "function",
  "function": {
    "name": "run_python",
    "description": "Execute Python 3.9 in a fresh interpreter starting in the persistent, networkless /work directory. Scientific packages and the indexed toolbox are importable; read-only case data is at /input/case.json. Return stdout, stderr, exit status, timeout status, and a citeable evidence_ref on success.",
    "parameters": {
      "type": "object",
      "properties": {
        "code": {"type": "string", "minLength": 1, "maxLength": 48000}
      },
      "required": ["code"],
      "additionalProperties": false
    }
  }
}
\end{Verbatim}

\begin{Verbatim}[breaklines=true,breakanywhere=true,fontsize=\scriptsize]
{
  "type": "function",
  "function": {
    "name": "submit_prediction",
    "description": "Submit the four scored glucose forecasts in horizon order.",
    "parameters": {
      "type": "object",
      "properties": {
        "target_timestamps_s": {
          "type": "array",
          "items": {"type": "integer"},
          "const": [1600622700, 1600624500, 1600626300, 1600628100]
        },
        "values_mgdl": {
          "type": "array",
          "items": {"type": "number"},
          "minItems": 4,
          "maxItems": 4
        },
        "reason": {"type": "string"},
        "evidence_refs": {
          "type": "array",
          "description": "Evidence refs from successful run_python results that directly support the public reason. Use [] if no Python call succeeded.",
          "items": {"type": "string", "minLength": 1},
          "maxItems": 14,
          "uniqueItems": true
        }
      },
      "required": ["target_timestamps_s", "values_mgdl", "reason", "evidence_refs"],
      "additionalProperties": false
    }
  }
}
\end{Verbatim}

Three constraints do real work.  The \texttt{const} on
\texttt{target\_timestamps\_s} pins the submission to the timestamps the
evaluation declares, so the agent cannot redefine what it is forecasting.
The \texttt{uniqueItems} flag prevents the same reference being cited
twice.  And \texttt{maxItems} matches the maximum number of programs, so
the citation list can never exceed the number of programs that could have
issued references.

\subsection{Optional helper modules}
\label{app:agent-helpers}

Two Python modules, 6.9~kilobytes of source between them, are copied into
the workspace before the first call, with two short description files, a
machine-readable index and a README; the bundle is 9.2~kilobytes.

\texttt{load\_packet()} parses \texttt{/input/case.json}, verifies the
outer and inner schemas, checks that horizons and target timestamps align,
and rejects any time-series interval, event or photograph record extending
beyond the exclusive cutoff.  \texttt{series\_frame()} returns one pandas
row per source interval and preserves nulls as pandas missing values.
\texttt{static\_pairs()} returns the literal baseline field/value pairs and,
by default, the derived HbA1c threshold category; passing
\texttt{include\_derived=False} returns only literal source fields.
\texttt{inventory()} reports per-signal coverage, missingness, last
observation time and recency; event counts by kind; photograph-record
counts and their recency; and literal and derived profile field counts.
None of these functions computes a forecast or declares a modality
relevant.

The catalog printed in the instruction is generated from the modules
rather than written by hand: the index supplies the module name, the
one-line relevance condition and the documentation path; the import line
and entry-point list are derived from the index; and the call signatures
are produced by parsing the modules' abstract syntax trees, so an
advertised signature cannot drift from the code.  The catalog is capped
at 4~kB and the bundle at 32 files and 256~kB.

\subsection{Sandbox}
\label{app:agent-sandbox}

Every origin gets a fresh sandbox that lives for the duration of the
episode and is then destroyed.  A program runs in a new interpreter inside
a \texttt{bubblewrap} container, a userspace Linux sandboxing utility,
with all namespaces unshared, including a fresh and empty network
namespace, so no network access of any kind is possible.  The container's
filesystem is built from scratch; Table~\ref{tab:agent-sandbox} lists
everything in it.

\begin{table}[htbp]
  \centering
  \small
  \begin{tabular}{p{0.24\linewidth}p{0.16\linewidth}p{0.50\linewidth}}
    \toprule
    Path & Access & Contents \\
    \midrule
    \texttt{/input/case.json} & read-only & the single cutoff-safe case file \\
    \texttt{/work}            & read-write & private scratch, seeded with the helper modules \\
    one further mount         & read-only & version-locked Python 3.9.25 with
                                            NumPy, SciPy, pandas,
                                            scikit-learn and statsmodels \\
    \texttt{/usr}, \texttt{/proc}, \texttt{/dev}, \texttt{/tmp} & read-only or virtual & base system runtime \\
    \bottomrule
  \end{tabular}
  \caption{The complete filesystem visible inside the sandbox.  Our source
  code, the user's home directory, the evaluation harness, the ground
  truth, credentials and all other forecast origins are absent.}
  \label{tab:agent-sandbox}
\end{table}

Resource limits are \sandboxtimeout{}~seconds of wall-clock time and
8~seconds of CPU time per program, \sandboxmem{}~MiB of address space, a
1~MiB per-file write limit, 64 open file descriptors, and no core dumps.
Environment variables are cleared and rebuilt with a fixed search path, a
fixed hash seed and single-threaded settings for the numerical libraries.
Captured standard output and standard error are limited to
\maxcapturebytes{} bytes each; if a program prints more, the head and tail
are retained around an inserted marker stating the original and omitted
byte counts, so the model is told that truncation happened and by how much.

\subsection{Control loop}
\label{app:agent-loop}

The loop is deterministic apart from the model's own sampling:

\begin{Verbatim}[breaklines=true,fontsize=\small]
open a fresh sandbox for this origin
messages <- [system instruction, per-origin task message]
python_calls <- 0
for turn in 1..16:
    offer run_python only if turn <= 14 and python_calls < 14
    reply <- model(messages, tools=offered, tool_choice="auto", max_tokens=6144)
    append reply to messages
    if reply contains no tool call:                      -> episode fails
    if reply contains submit_prediction:
        it must be the only call in that reply           -> otherwise fails
        validate it against the rules given below
        if valid:   return the values
        else:       append the validation error as the tool result; continue
    for each run_python call in the reply:
        if python_calls >= 14: append a budget-exhausted error; continue
        python_calls <- python_calls + 1
        result <- execute in the sandbox
        if the program exited 0 and did not time out:
            attach a fresh evidence reference to the result
        append the result as the tool result
episode fails if the budget is exhausted with no valid submission
\end{Verbatim}

An episode that fails produces no forecast; the harness records the failure
rather than substituting a fallback.

Two mechanics are worth stating.  The model may issue several programs in
one reply, and does so in the opening turn of most episodes, so the two
budgets are not consumed in lock-step.  And withholding the Python action
in the last two model calls is a guard against an episode that would
otherwise let its budget lapse without submitting; a typical episode stops
well short of either budget and never reaches it.

\subsection{Evidence references and submission validation}
\label{app:agent-evidence}

When a program exits successfully the harness computes an \emph{evidence
reference} of the form \texttt{python:$k$:sha256:$h$}, where $k$ indexes the
program within the episode and $h$ is the SHA-256 digest of a canonical
JSON encoding of three things together: the case identifier, cutoff and
target timestamps; the exact program source; and the exact result object
including its captured output.  The reference is appended to the tool
result the model sees, and the harness retains the set of references it has
issued in this episode.

A submission is rejected if the argument set does not match the schema; if
the target timestamps are not exactly those of this case; if the values are
not finite numbers, one per requested timestamp; if the reason is
empty; if the citation list contains a duplicate, a non-string or more than
\maxcodecalls{} entries; if it cites a reference not in the issued set; or
if a program succeeded but nothing is cited.  The rejection is returned to
the model as the result of its own action, and it may correct and resubmit
while calls remain.  Across the 576 origins, 39 submissions from 24 origins
were rejected, 38 of them for citing a reference that had never been issued
and one for citing the same reference twice; all 24 origins subsequently
submitted successfully.

Two distinct properties hold and are worth separating.  A fabricated
citation is caught by the set-membership check, not by the digest.  What
the digest adds is that a reference cannot be transplanted from a different
origin or attached to a different program, because it is computed over the
program, its output and the case identity together.

\subsection{Leakage controls}
\label{app:agent-leakage}

Two mechanisms keep post-cutoff information out, and they are independent.

First, the case file is built to be cutoff-safe and re-checked when read:
the accessor refuses to serve an interval whose exclusive end is later than
the cutoff, or an event or photograph record whose timestamp is later than
the cutoff.  The helper module re-validates the same conditions when the
agent loads the file, but that second check is voluntary, since the agent
may read the file directly, so it is a convenience rather than a guarantee.

Second, the sandbox contains nothing else.  There is no ground truth, no
other forecast origin, no evaluation code, no host filesystem and no
network, so there is no channel through which post-cutoff information could
arrive even if the agent looked for it.  Scoring happens outside the
sandbox against a separate ground-truth store no episode can reach, and the
agent's output is pinned to the declared target timestamps.

\subsection{Evaluation guarantees}
\label{app:evaluation-guarantees}

Four properties of the evaluation are what make the comparisons in this
paper meaningful.  We state them together here because they are argued
separately above and in the evaluation appendices.

First, cutoff safety.  Nothing recorded at or after the forecast cutoff
can reach a method.  For \agentname{} this is enforced twice and
independently, as Section~\ref{app:agent-leakage} describes: the case
file is built and re-checked to contain nothing past the cutoff, and the
sandbox contains no other source of information.

Second, a frozen method.  The configuration in this appendix was
hash-pinned before any held-out origin was scored and was not changed
afterwards.  We did not select a configuration by looking at held-out
scores.

Third, identical cases and identical truth.  Every method, ours and every
baseline, is scored on the same forecast origins with the same inputs
available before the cutoff, against the same stored truth
(Sections~\ref{sec:data-preparation-subsampling}
and~\ref{sec:tsfm-baseline-setup}).  The origin sets are recorded as
versioned manifests, so what was scored is a fact on disk rather than a
description.

Fourth, one scorer.  Predictions are written per origin and scored
afterwards by the same code for every method, using the metric the
benchmark defines.  Scoring happens outside the sandbox, against a truth
store that no episode can reach.

\subsection{Glucose-only configuration}
\label{app:agent-glucose-only}

The glucose-only configuration shares the task sentence, the instruction
structure, the model, the reasoning effort, the budgets, the sandbox and
the control loop.  Three things differ.

The case object is restricted to a schema with exactly six keys: a schema
tag, an opaque identifier, the exclusive cutoff, the sampling interval in
minutes, a timestamp array and a glucose array.  The enclosing object adds
the forecast offsets, the scored horizon labels and the target timestamps.
The loader rejects any unexpected key, so no other signal, no baseline
profile and no participant identifier can be added silently.  The helper
modules shrink to match: the table accessor returns timestamps and glucose
only, the availability report covers glucose only, and the profile accessor
does not exist.

The submission action accepts exactly as many values as the case declares,
which lets one agent produce four point forecasts on CGMacros and
GlucoFM-Bench and a 24-step five-minute trajectory on the CGM-LSM OhioT1DM
protocol with no benchmark-specific prompt or code branch.

The instruction differs only where the catalog is regenerated from the
reduced modules, where four catalog lines change and one is removed,
giving 2{,}125 characters against 2{,}232, and where the two action
descriptions mention glucose-only data.  The executor paragraph, the task
sentence, and the evidence and submission paragraph are byte-identical.

\subsection{Identity and reproduction}
\label{app:agent-identity}

The first 576 of the 1{,}903 held-out origins were run in four batches and
the remaining 1{,}327 in a fifth, and all were scored cumulatively.  Each
batch used the same command,
changing only the list of origins and the run identifier: the harness takes
the agent implementation, the evaluation set, the batch's origin list, the
model, the reasoning effort and a per-origin cap of \maxturns{} model
calls, and running the held-out set requires an explicit opt-in flag.
Origins within a batch are processed concurrently, which affects wall-clock
time only, since each origin is an independent episode.

For anyone reproducing the run: over the 576 audited origins, a typical
origin used about 7 model calls
and 7 Python programs, roughly 95{,}000 cumulative input tokens and
6{,}400 output tokens, and 174 seconds of wall-clock time, all medians.
Token counts are cumulative because the whole conversation is resent at
each model call.

\clearpage

\section{Design Variants}
\label{sec:design-variants}

Two variants of \agentname{} were run on the same 1{,}903 held-out CGMacros
origins with everything else frozen: one that reaches the context through a
hand-designed menu of retrieval tools instead of a Python environment,
and the same agent driven by other language models and reasoning settings.

\subsection{An Alternative Hand-Designed Tool Interface}
\label{app:hand-design-comparison}

The design choice examined here is the shape of the data interface.  Our
method gives the agent the complete cutoff-safe context as one file plus a
Python environment.  The natural alternative is to hand-design
one retrieval tool per access pattern and let the agent call them.  We
built that alternative first, and this appendix describes it and reports
how the two compare on the same origins.

\subsubsection{The interface}

The hand-designed interface offers six retrieval tools plus a terminal
submission tool, listed in Table~\ref{tab:hand-designed-tools}.  Every tool
returns structured JSON and issues an evidence reference that the
submission must cite.  The four tools that read the context draw on the
same eight signals our case file exposes.

\begin{table}[htbp]
  \centering
  \small
  \begin{tabular}{p{0.26\linewidth}p{0.66\linewidth}}
    \toprule
    Tool & What it returns \\
    \midrule
    read a time series &
      the values of chosen signals over a chosen interval, paginated \\
    summarize an interval &
      deterministic summary statistics for chosen signals over a chosen
      interval \\
    list events &
      the recorded events in a chosen interval, optionally filtered by
      kind, paginated \\
    read the interval around an event &
      chosen signals for a chosen number of minutes before and after a
      chosen event, paginated \\
    list meal photographs &
      metadata for the photographs available before the cutoff, optionally
      restricted to an interval or one event, paginated \\
    attach a meal photograph &
      the pixels of one chosen photograph, inserted into the model's
      context \\
    submit the forecast &
      terminal; the forecast values and a structured justification \\
    \bottomrule
  \end{tabular}
  \caption{The hand-designed interface.  Every access pattern available to
  the agent is one of these seven calls.}
  \label{tab:hand-designed-tools}
\end{table}

The submission is more structured than ours: it carries the forecast
values, a summary sentence, and a list of factors, each with a claim, a
direction from a fixed set (\emph{up}, \emph{down}, \emph{neutral},
\emph{uncertain}) and at least one evidence reference, plus optional stated
uncertainties and a confidence in $[0,1]$.

\subsubsection{Instruction}

The system instruction is short and concerns evidence discipline rather
than analysis.  The per-origin message, by contrast, directs the analysis:
it names recent trajectories, personal meal and activity timing patterns,
prior meal analogs and same-time activity patterns.  The complete rendered
system instruction and both user-content blocks from the reported
faithful-reproduction run appear, for completeness, in
Section~\ref{app:faithful-rendered-prompt}.  An ``\texttt{e:} reference'' is
that interface's name for an evidence reference, playing the same role as
the \texttt{python:} references of Section~\ref{app:agent-evidence}.

This is the structural difference between the two designs.  Here the
instruction must name specific tools for specific situations, because the
tool set is the analysis vocabulary: an operation the tools do not express
is unavailable however useful it would be at a particular origin.  In our
interface the equivalent sentence is ``Choose your own analysis.''

\subsubsection{What is held fixed}

Table~\ref{tab:interface-differences} lists the differences.  Both designs
run on the same 1,903 held-out origins, with the same language model, the
same reasoning setting, the same input files, the same ground truth and
the same scorer.

Two differences are not controlled, and neither favors our method.  The
nominal per-origin budgets differ, because each design was frozen with the
budget it was built for, but the budget did not bind: over the 576 origins
whose full traces we audited, the hand-designed agent used a median of 4
model calls and never more than 6 of its 7, so it always stopped of its own
accord rather than being cut off.  And the hand-designed interface can
attach meal photographs as image input, whereas our agent receives their
timestamps only.  The comparison is therefore between two complete designs
rather than between one factor and its absence.

\begin{table}[htbp]
  \centering
  \small
  \begin{tabular}{p{0.29\linewidth}p{0.28\linewidth}p{0.33\linewidth}}
    \toprule
                                & \agentname{} & Hand-designed \\
    \midrule
    Data access                 & one read-only JSON file & six retrieval tools \\
    Computation                 & arbitrary Python code & none beyond the tools \\
    Method guidance in the instruction & none & names tools and analyses \\
    Model calls per origin      & at most \maxturns{} & at most 7 \\
    Rounds that may call tools  & at most \maxcodecalls{} & at most 6 \\
    Non-terminal calls          & at most \maxcodecalls{} Python programs
                                & at most 24 retrieval calls \\
    Generated tokens per call   & \maxouttokens{} & 8{,}192 \\
    Meal photographs            & timestamps only & pixels, on request \\
    Forecast output             & 4 values, all scored
                                & 24 values on a 5-minute grid, of which 4
                                  are scored \\
    Justification format        & free text plus citations
                                & structured factors, each with a direction
                                  and citations \\
    \midrule
    Language model              & \multicolumn{2}{l}{identical, at the same high reasoning setting} \\
    Origins, inputs, truth, scorer & \multicolumn{2}{l}{identical} \\
    \bottomrule
  \end{tabular}
  \caption{Differences between the two interfaces.  The lower block lists
  what is held fixed.}
  \label{tab:interface-differences}
\end{table}

\subsubsection{Result}

Table~\ref{tab:appendix-hand-design-multimodal} reports both designs, at
two reasoning settings for the hand-designed one, together with the
multimodal time series foundation model baseline, on the same 1,903
held-out origins.  The Python environment is the more accurate
of the two designs at all four horizons, and by 1.71~mg/dL on the mean
over them.  We read this as evidence that the fixed tool menu was
limiting rather than merely redundant, while noting the two uncontrolled
differences above.

One caveat matters more than the rest, and it is a limit on what this
comparison can support.  We compared our design against \emph{one}
hand-designed interface, the one we built first.  A different tool set,
or a different instruction over the same tools, could do better.  This
result is therefore necessary evidence for the choice we made and not
sufficient evidence that hand-designed interfaces are the worse idea in
general.

\begin{table}[htbp]
  \caption{Comparison with the hand-designed tool interface on 1,903 exactly matched
  CGMacros HOLDOUT origins. Cells report participant-level RMSE mean (sample
  standard deviation), in mg/dL. Mean-4 is computed per participant across
  the four horizons before aggregation.}
  \label{tab:appendix-hand-design-multimodal}
  \centering
  \scriptsize
  \setlength{\tabcolsep}{3.0pt}
  \resizebox{\textwidth}{!}{%
  \begin{tabular}{lrrrrr}
    \toprule
    Method & 30 min & 60 min & 90 min & 120 min & Mean-4 \\
    \midrule
    Hand-designed faithful port (high effort) & \underline{17.193} (3.559) & 23.985 (8.333) & 27.186 (10.737) & 28.575 (12.852) & 24.235 (8.730) \\
    Hand-designed faithful port (medium effort) & 17.897 (4.202) & 25.631 (9.483) & 28.762 (12.085) & 30.258 (14.567) & 25.637 (9.842) \\
    Chronos-2 (fuller multimodal) & 17.246 (5.125) & \underline{23.711} (9.586) & \underline{26.052} (11.527) & \underline{26.866} (12.887) & \underline{23.469} (9.658) \\
    \agentname{} (agent-designed multimodal) & \textbf{16.357} (3.605) & \textbf{22.132} (8.220) & \textbf{25.163} (9.663) & \textbf{26.450} (11.333) & \textbf{22.526} (8.078) \\
    \bottomrule
  \end{tabular}}
  \vspace{2pt}

  {\scriptsize \textbf{Bold} marks the lowest RMSE and \underline{underlining} the second-lowest in each column; only the mean is styled, not the standard deviation. Rankings use unrounded values.}
\end{table}

\subsubsection{Complete rendered prompt for the faithful-reproduction manual design}
\label{app:faithful-rendered-prompt}

For completeness, we reproduce below the complete message content of one
first request from the faithful-reproduction design used in the comparison
above.  Its user content is represented on the wire as two consecutive text
blocks, which are shown separately rather than concatenated.  The action
schemas for the seven actions summarized in
Table~\ref{tab:hand-designed-tools} are supplied alongside these messages.

\paragraph{System instruction}

\begin{Verbatim}[breaklines=true,breakanywhere=true,fontsize=\scriptsize]
Forecast glucose for this autonomous CGMacros episode using only cutoff-safe
episode facts supplied in the request or returned by its tools. Do not use
post-cutoff observations, outcomes, future actions, or unprovided person facts.
General knowledge is not episode evidence. Do not expose private chain-of-thought.
Submit exactly one terminal prediction with a concise public reason. Every reason
factor must cite at least one supplied runtime-issued e: reference; no other
identifier is an evidence citation.
\end{Verbatim}

\paragraph{Per-origin user content}

The first text block renders the origin metadata, complete untimed baseline
profile, forecast-grid contract, evidence rules and analysis guidance.

\begin{Verbatim}[breaklines=true,breakanywhere=true,fontsize=\scriptsize]
This is a regular fixed-schedule forecast origin for subject 009, independent of
whether a meal occurred. numeric_history=[1600160700,1600621200); cutoff_exclusive_s=1600621200.
Ordinary point events and historical images may be visible through 1600621200. Timestamps
are date-shifted dataset seconds with no timezone; do not infer calendar, weekday,
season, or clinical date. health_context: study_baseline_profile is explicitly granted untimed study-baseline raw text; measurement_time_s=not_provided. Each [field,value] pair is literal source text and an empty value is a source empty. Units exist only where the field or value literally supplies them. Do not infer timing, units, diabetes type, medications, diagnoses, or medical records. This context may condition point values and may be cited only through the run-local e: reference supplied alongside this profile.
study_baseline_profile_source_id=static-profile:c70dcbb03584bb484dff71d91390a7b120f84b421f76284e5216845e846f9ab3
study_baseline_profile=[["Age","34"],["Gender","F"],["BMI","37.00150587"],["Body weight","183.2"],["Height","59"],["Self-identify","Hispanic/Latino"],["A1c PDL (Lab)","5.7"],["Fasting GLU - PDL (Lab)","122"],["Insulin","25.7"],["Triglycerides","312"],["Cholesterol","202"],["HDL","40"],["Non HDL","162"],["LDL (Cal)","108"],["VLDL (Cal)","62"],["Cho/HDL Ratio","5.1"],["Collection time PDL (Lab)","7:32:00 AM"],["#1 Contour Fingerstick GLU","119"],["Time (t)","10:29"],["#2 Contour Fingerstick GLU","128"],["#3 Contour Fingerstick GLU","96"]]
derived_context: A1c PDL (Lab) is interpreted on the numeric A1c percent scale. The ADA-threshold category is normal-range when A1c <5.7, prediabetes-range when A1c >=5.7 and <6.5, and diabetes-range when A1c >=6.5. It is untimed baseline context, not a diagnosis or diabetes type, and shares the profile's run-local e: reference.
study_baseline_derived_fields=[["A1c category (ADA thresholds, derived)","prediabetes-range"]]
No memory, replay lesson, retrieval state, or current image is supplied. Do not
infer diabetes type, medications, diagnoses, medical records, or person facts not
explicitly present in the supplied baseline profile.

Return exactly 24 finite point_mg_dl values in mg/dL for the terminal tool's fixed
five-minute grid. Its first timestamp is the first unobserved bin at 1600621200; subsequent
timestamps are 300 seconds apart.

Allowed episode evidence is supplied glucose, heart rate, METs, activity energy,
meal macros/events, and offered-tool results. Historical images are optional and
must be explicitly discovered and attached with the offered tools. An absent,
unpresented, or failed-to-attach image has unknown contents: do not impute it, treat
it as negative evidence, or cite it. Do not assume a future meal or action.

Gather evidence before predicting, using the available tools selectively when their
expected information gain is relevant. Inspect recent trajectories and personal
meal/activity timing patterns. When a forecast-relevant claim relies on a prior
meal analog, inspect its observed glucose and activity response with
get_event_interval rather than infer that response from meal attributes alone. Use
read_time_series or summarize_interval for same-time activity patterns. Treat
sparse analog evidence as sparse rather than forcing a match.
\end{Verbatim}

The second text block supplies the run-local evidence reference for the
baseline profile rendered in the first block.

\begin{Verbatim}[breaklines=true,breakanywhere=true,fontsize=\scriptsize]
Static baseline profile evidence: e:95d2d138e2:meal-onset-lite-static-baseline-v1:1. Cite this reference for claims grounded in study_baseline_profile.
\end{Verbatim}

\subsection{Which language model, and how much reasoning}
\label{sec:additional-llm-ablation}

Table~\ref{tab:appendix-llm-ladder} repeats the three input settings of
Section~\ref{sec:results-cgmacros} with the same agent driven by the same
model at a lower reasoning setting and by a smaller model of the same
family.  Every row gains from the longer history, and the gain from the
non-glucose signals is 1.4, 1.8 and 0.3~mg/dL on the mean over horizons for
the three rows, so the smaller model gains as little from them as Chronos-2
does.  The stronger the model the lower the error in every setting, by 1.2
to 2.1~mg/dL for the reasoning setting and 3.0 to 5.2 for the model size.
The two frontier models from other families, run in the multimodal
setting only on the same held-out origins with everything else frozen,
are in Table~\ref{tab:llm-multimodal-ablation} in the main text; they
finish 1.1 and 3.6~mg/dL behind ours on the mean over horizons.
Table~\ref{tab:llm-claude-cgm-ablation} is the ladder with standard
deviations.

\begin{table}[htbp]
  \caption{\textbf{The same agent driven by three language models and reasoning settings.}  Participant-level RMSE (mg/dL) in the three input settings.  Mean is over the four horizons.  \textbf{Bold} marks the lowest error within a setting.  Every row gains from the longer history, and the gain from the non-glucose signals is 1.4, 1.8 and 0.3~mg/dL on the mean for the three rows.  Standard deviations are in Table~\ref{tab:llm-claude-cgm-ablation} and two models from other families in Table~\ref{tab:llm-multimodal-ablation}.}
  \label{tab:appendix-llm-ladder}
  \centering\small
  \begin{tabular}{llrrrrr}
    \toprule
    Input setting & Language model (effort) & 30 & 60 & 90 & 120 & Mean \\
    \midrule
    \multirow{3}{*}{Glucose only, 12 h} & Claude Opus 5 (high), ours & \best{18.68} & \best{27.21} & \best{31.08} & \best{33.19} & \best{27.54} \\
     & Claude Opus 5 (medium) & 19.59 & 28.73 & 32.32 & 34.15 & 28.70 \\
     & Claude Sonnet 5 (high) & 19.59 & 29.39 & 34.62 & 38.72 & 30.58 \\
    \addlinespace
    \multirow{3}{*}{Glucose only, full history} & Claude Opus 5 (high), ours & \best{17.42} & \best{24.19} & \best{26.58} & \best{27.66} & \best{23.96} \\
     & Claude Opus 5 (medium) & 18.62 & 26.24 & 29.04 & 30.33 & 26.06 \\
     & Claude Sonnet 5 (high) & 19.03 & 27.12 & 31.64 & 34.36 & 28.04 \\
    \addlinespace
    \multirow{3}{*}{Full multimodal context} & Claude Opus 5 (high), ours & \best{16.36} & \best{22.13} & \best{25.16} & \best{26.45} & \best{22.53} \\
     & Claude Opus 5 (medium) & 17.02 & 23.75 & 27.31 & 29.00 & 24.27 \\
     & Claude Sonnet 5 (high) & 18.59 & 26.61 & 31.11 & 34.51 & 27.71 \\
    \bottomrule
  \end{tabular}
\end{table}

\begin{table}[htbp]
  \caption{Claude-family ablation across matched input settings on
  1,903 CGMacros HOLDOUT origins. Cells report participant-level RMSE mean
  (sample standard deviation), in mg/dL. Mean-4 is computed per participant
  across the four horizons before aggregation.}
  \label{tab:llm-claude-cgm-ablation}
  \centering
  \scriptsize
  \setlength{\tabcolsep}{2.4pt}
  \resizebox{\textwidth}{!}{%
  \begin{tabular}{lllrrrrr}
    \toprule
    Input setting & Model & Effort & 30 min & 60 min & 90 min & 120 min & Mean-4 \\
    \midrule
    \multirow{3}{*}{Multimodal} & Claude Opus 5 (ours) & high & \textbf{16.357} (3.605) & \textbf{22.132} (8.220) & \textbf{25.163} (9.663) & \textbf{26.450} (11.333) & \textbf{22.526} (8.078) \\
     & Claude Opus 5 & medium & \underline{17.016} (3.683) & \underline{23.746} (8.594) & \underline{27.308} (10.659) & \underline{28.999} (12.550) & \underline{24.267} (8.711) \\
     & Claude Sonnet 5 & high & 18.595 (4.177) & 26.613 (8.857) & 31.108 (11.528) & 34.511 (14.200) & 27.707 (9.486) \\
    \addlinespace
    \multirow{3}{*}{CGM only (full history)} & Claude Opus 5 (ours) & high & \textbf{17.416} (4.930) & \textbf{24.187} (9.150) & \textbf{26.584} (10.846) & \textbf{27.661} (12.395) & \textbf{23.962} (9.199) \\
     & Claude Opus 5 & medium & \underline{18.624} (4.815) & \underline{26.240} (9.921) & \underline{29.042} (12.744) & \underline{30.328} (14.391) & \underline{26.058} (10.348) \\
     & Claude Sonnet 5 & high & 19.026 (5.260) & 27.124 (9.915) & 31.639 (12.622) & 34.358 (14.623) & 28.037 (10.492) \\
    \addlinespace
    \multirow{3}{*}{CGM only (12 h)} & Claude Opus 5 (ours) & high & \textbf{18.679} (5.080) & \textbf{27.209} (10.093) & \textbf{31.077} (12.579) & \textbf{33.191} (14.883) & \textbf{27.539} (10.496) \\
     & Claude Opus 5 & medium & \underline{19.588} (4.919) & \underline{28.731} (9.956) & \underline{32.320} (12.922) & \underline{34.151} (15.511) & \underline{28.698} (10.646) \\
     & Claude Sonnet 5 & high & 19.591 (5.789) & 29.392 (11.039) & 34.620 (13.951) & 38.720 (16.741) & 30.581 (11.699) \\
    \bottomrule
  \end{tabular}}
  \vspace{2pt}

  {\scriptsize \textbf{Bold} is best within each input setting and \underline{underlining} is second best;
  only means are styled. Rankings use unrounded participant-macro RMSE values.}
\end{table}

\clearpage
\section{Data Preparation and Subsampling}
\label{sec:data-preparation-subsampling}

Scoring \agentname{} on every eligible forecast origin requires several
language-model calls per origin, so on GlucoFM-Bench and on the CGM-LSM
OhioT1DM benchmark we score every method on one fixed subset, built to keep
the benchmark's weighting and its main sources of variation.  On CGMacros
the accuracy results use the full held-out set of 1,903 origins, and a
576-origin subsample of it carries the behavior analyses, which read the
full trace of every episode.  The full populations remain the reference
evaluations, and a subset score is compared only with methods scored on
exactly the same origins.

\paragraph{Common design principles.}
All three samplers use the same seed, \texttt{20260809}, fixed in code before
any agent run, and select on nothing that depends on the outcome: the
dataset, the participant, the forecast time, the rank of the available
history and, on CGMacros, the meals and activity recorded before the cutoff.
Glucose targets, model predictions, errors and scores never enter the
selection.

\begin{table*}[t]
  \centering
  \small
  \setlength{\tabcolsep}{4pt}
  \begin{tabular}{p{0.20\linewidth}p{0.12\linewidth}p{0.28\linewidth}p{0.30\linewidth}}
    \hline
    Benchmark & Eligible full size & Evaluated origins & Preserved weighting structure \\
    \hline
    CGMacros & 1,903 & all 1,903; a 576-origin subsample for the behavior analyses & Equal number of origins per participant; participant-specific situation distribution \\
    GlucoFM-Bench & 495,917 & 8,646 & At most 32 origins per participant within each dataset \\
    CGM-LSM OhioT1DM & 9,214 & 1,536 & Full-population participant proportions \\
    \hline
  \end{tabular}
  \caption{Full populations and the origins the paper scores.}
  \label{tab:subsampling-summary}
\end{table*}

\paragraph{CGMacros.}
The held-out CGMacros evaluation set contains 1,903 forecast origins from
nine participants, set aside by a seeded random split of the 45 participants
(20\% held out) that was frozen before any model development.  They span
the three ADA A1c ranges evenly, three each, and all nine self-identify as
Hispanic/Latino, as do 34 of the 45
(Table~\ref{tab:holdout-participants}).  Over the 1,903 origins the median
origin has 131 hours (5.5 days) of history and 22 logged meals before its
cutoff (interquartile ranges 78 to 184 hours and 13 to 31 meals).  Forecast
origins are spaced one hour apart, while inputs and targets are represented
on a five-minute grid.

\begin{table}[htbp]
  \centering
  \small
  \setlength{\tabcolsep}{4pt}
  \begin{tabular}{@{}lrcrrlrrrr@{}}
    \toprule
    & & & & \multicolumn{2}{c}{A1c} & Fasting & Duration & CGM & Logged \\
    \cmidrule(lr){5-6}
    Participant & Age & Sex & BMI & (\%) & range & (mg/dL) & (days) & readings & meals \\
    \midrule
    001 & 27 & M & 22.3 & 5.4 & Normal       &  91 & 10.2 & 14{,}265 & 43 \\
    003 & 59 & F & 26.9 & 6.5 & Diabetes     & 118 & 10.1 & 14{,}260 & 35 \\
    004 & 33 & F & 42.4 & 5.5 & Normal       & 105 &  9.9 & 14{,}275 & 62 \\
    009 & 34 & F & 37.0 & 5.7 & Prediabetes  & 122 & 10.0 & 14{,}226 & 41 \\
    010 & 54 & F & 35.8 & 5.7 & Prediabetes  & 100 & 11.2 & 14{,}275 & 45 \\
    013 & 56 & F & 32.6 & 6.2 & Prediabetes  & 103 & 11.0 & 14{,}275 & 39 \\
    033 & 40 & M & 30.0 & 5.6 & Normal       & 116 & 11.0 & 14{,}215 & 33 \\
    038 & 52 & M & 34.5 & 7.2 & Diabetes     & 151 & 10.1 & 14{,}270 & 40 \\
    039 & 45 & F & 26.9 & 8.3 & Diabetes     & 158 & 12.0 & 14{,}275 & 34 \\
    \bottomrule
  \end{tabular}
  \caption{The nine held-out CGMacros participants.  Baseline values are the
  study-entry laboratory panel released with the dataset; the A1c range
  applies the ADA thresholds used for the agent's profile field (normal
  $<5.7$, prediabetes $5.7$ to $<6.5$, diabetes $\geq 6.5$).  CGM readings
  are Dexcom G6 Pro samples.  Each participant contributes 208 to 212 of the
  1{,}903 forecast origins.}
  \label{tab:holdout-participants}
\end{table}

On CGMacros we score every method on all 1,903 origins.  The behavior
analyses of Appendix~\ref{sec:behavior-appendix} instead read the full trace
of every episode, so for them we audit a subsample of 576 origins.  Because
the CGMacros metric weights participants equally, we take 64 origins from
each of the nine participants.  Within a participant
we first divide the origins into the four situations used throughout the
paper, by what the available context shows shortly before the cutoff.
A meal score sums the carbohydrate of earlier meals, capped at 150~g per
meal and discounted exponentially with a 90-minute time constant, and an
origin counts as near a meal when the score is at least 10~g.  An activity
score sums the minutes of earlier moderate-activity bouts, discounted from
the end of each bout with a 60-minute time constant, and counts as near
exercise at 5~minutes or more; a bout is at least five minutes at 3.0
metabolic equivalents (METs) or above, or at vendor activity intensity 2 or
above where METs are missing, and is truncated at the cutoff if it crosses
it.  An origin is \emph{post-meal} when only the meal score is active,
\emph{post-exercise} when only the activity score is active, \emph{meal +
exercise} when both are, and \emph{quiet} when neither is.  Both scores
read only events strictly before the cutoff, the label was used only for
sampling, and no forecaster ever received it.

Each participant's quota of 64 is then split across the four situations in
the proportions of that participant's full set, by Hamilton's
largest-remainder method, which rounds proportional shares to integers that
sum to the quota.  Within a situation we order the origins by the rank of
their available history and take them by systematic sampling from a random
offset, which spreads the selected origins across the available history instead of
concentrating them in one stretch.

\paragraph{GlucoFM-Bench.}
On GlucoFM-Bench the eligible population is every origin that persistence,
Chronos~2 and TimesFM~2.5 all support at all four horizons (15, 30, 60 and
90 minutes): 495,917 origins from 273 participants across nine datasets.
CGMacros is evaluated separately.  GlucoFM-Bench averages participants within each dataset,
so we sample within each dataset--participant pair.  A seeded hash of the
dataset, the participant and the forecast cutoff gives every eligible origin
a priority, and the priorities order a participant's origins as a
reproducible random permutation that consults no outcome.  The evaluated
subset keeps the first 32 origins of each participant, 8,646 origins in all,
fewer than the 8,736 that 32 origins from every participant would give
because a few participants have fewer than 32 eligible origins.

\paragraph{CGM-LSM OhioT1DM.}
The CGM-LSM OhioT1DM evaluation set contains 9,214 forecast instances from
11 participants, and its primary metric is the unweighted mean over
instances rather than over participants, so giving every participant the
same number of instances would change what the benchmark measures.  The
evaluated subset holds 1,536 instances, allocated across participants in
proportion to their share of the 9,214 by the same largest-remainder
method.  Within a participant the sampler orders the
instances in time and takes them by seeded systematic sampling from a random
offset, which keeps each participant's share and spreads the instances over
their observation period.

\paragraph{Reproducibility and fair comparison.}
Each selection is recorded in a versioned manifest with the selected origin
identities, the allocation counts, the seed and the source hashes, and
automated checks verify the exact sizes, deterministic reconstruction, and
the absence of target and error fields from the selection inputs.  Every
agent and baseline is scored on the identical selected identities; the
foundation-model baselines store a prediction for every selected origin in
the evaluation pipeline, and previously saved forecasts serve only to check
reproduction fidelity where a reference implementation exists.  We tuned
nothing on the selected origins, and we compare a subset score only with
methods scored on exactly the same origins.

\clearpage
\section{Time Series Foundation Model Baselines}
\label{sec:tsfm-baseline-setup}

Every time series foundation model (TSFM) baseline is run zero-shot from a
fixed pretrained checkpoint on the evaluated origins of
Appendix~\ref{sec:data-preparation-subsampling}, with no fitting or tuning
on the evaluation participants and with nothing recorded after the forecast
cutoff as input.

\paragraph{GlucoFM-Bench.}
On GlucoFM-Bench we follow the benchmark codebase's preprocessing, model
interfaces and evaluation protocol: each model receives 12 hours of glucose
history (144 five-minute samples) and predicts the 15-, 30-, 60- and
90-minute horizons, and errors are pooled within a participant and then
averaged across participants within each dataset.  The models are
persistence, Chronos~2, TimesFM~2.5, Moirai~2.0 and Timer, each with its
reference zero-shot point forecast.

\paragraph{CGMacros glucose-only baselines.}
On CGMacros the glucose-only baselines are persistence, which repeats the
last glucose value, and Chronos~2, TimesFM~2.5, Moirai~2.0 and Timer with
glucose as the only input, run once with the final 12 hours of history and
once with the whole pre-cutoff glucose history, predicting 30, 60, 90 and
120 minutes ahead (all four appear in Table~\ref{tab:main-cgmacros-holdout}).
Timer preserves its upstream 96-step patch flooring, so its 12-hour
setting sees 8 hours of history, and uses causal-mean imputation for
missing historical CGM; Table~\ref{tab:main-cgmacros-holdout} marks its
rows with a dagger for that reason.

\paragraph{CGMacros multimodal Chronos~2.}
CGMacros has no standard multimodal TSFM configuration, so we build one
from Chronos~2's native covariate interface and everything the dataset
records.  Alongside glucose the model receives, as aligned past covariates,
heart rate; activity energy expenditure and metabolic equivalents (METs);
the meal count; indicators for breakfast, lunch, dinner, snack and other;
meal energy, carbohydrate, protein, fat, fiber and amount consumed; and,
where a numeric signal can be missing, an indicator of whether it was
observed.  Sinusoidal time-of-day and day-of-week features are supplied
into the forecast horizon as well, because the calendar is known in
advance, while every physiological, activity and meal variable stops
strictly at the cutoff.  All covariates share the causal five-minute grid
of the glucose series.

\paragraph{CGM-LSM OhioT1DM.}
On OhioT1DM we follow the CGM-LSM protocol: each instance carries a
24-hour glucose history and asks for a 24-step forecast at five-minute
resolution, and persistence and the four foundation models are scored by
the benchmark evaluator as cumulative RMSE over the first 30, 60 and 120
minutes (Timer takes the last 288 of the 289 context points, since its
input patch is 96 steps).

\paragraph{Reproducibility.}
All baseline predictions pass through the same evaluation pipeline as our
method, and each run records the benchmark subset, the checkpoint and
software configuration, the per-origin predictions and the resulting score.
Where a reference benchmark implementation exists we follow it; on CGMacros
the multimodal configuration uses the model's supported covariate interface
with the causally available signals listed above.

\clearpage

\section{CGMacros in Detail}
\label{sec:additional-results}

This appendix gives the CGMacros comparison of Section~\ref{sec:results-cgmacros}
in full: the same result within each situation, and the same result at
each horizon.

\subsection{CGMacros results within each situation}
\label{sec:additional-situations}

The results figure summarizes the situation breakdown.  The four tables
below give it in full: the same 1,903 held-out origins as the main
CGMacros table, restricted in turn to the origins labeled quiet,
post-meal, post-exercise, and meal + exercise, with every method and
every input setting.

We hypothesize that Chronos-2 gains after exercise and \agentname{} after meals because the two
kinds of information reach a forecaster in different forms.  Exercise
arrives as heart-rate and activity time series sampled alongside glucose,
the covariate a time series foundation model is built to take next to the
target, while a meal arrives as a single event, a timestamp with a label
such as dinner and a few nutrient numbers, which as a covariate series is
empty except at one point, far from the continuous series such a model was
pretrained on.  The nutrient numbers do carry signal, since a model
pretrained on glucose learns to use them once a head is trained on labeled
meals~\citep{glucofm}, but \agentname{} needs no such training: it uses the
meal as an event, aligning the person's past responses on it and reading or fitting a
response from them, as the cases in
Section~\ref{sec:results-behaviour} show.

\begin{table}[htbp]
  \caption{\textbf{CGMacros held-out forecasting in the quiet situation.} The same 1,903 held-out origins as Table~\ref{tab:main-cgmacros-holdout}, restricted to the 917 origins labeled quiet. Situation labels were attached after every forecast was scored; no forecaster saw them and no origin was selected or dropped on them. Within each participant, RMSE uses only that participant's origins in this situation; cells report the unweighted participant mean (sample standard deviation), in mg/dL.}
  \label{tab:cgmacros-holdout-stratum-quiet}
  \centering
  \small
  \setlength{\tabcolsep}{4pt}
  \begin{tabular}{lrrrr}
    \toprule
    Method & 30 min & 60 min & 90 min & 120 min \\
    \midrule
    \multicolumn{5}{l}{\textit{Reference}} \\
    \quad Persistence & 12.458 (4.522) & 19.688 (8.463) & 27.021 (10.482) & 31.142 (13.890) \\
    \addlinespace
    \multicolumn{5}{l}{\textit{CGM only, 12 hours of history}} \\
    \quad TimesFM-2.5 & \best{11.251} (4.264) & \best{18.031} (7.723) & \best{24.303} (9.373) & \best{27.571} (12.566) \\
    \quad Moirai-2.0 & 11.846 (4.353) & 19.066 (7.843) & 25.559 (9.902) & 28.944 (13.196) \\
    \quad Timer$^{\dagger}$ & 12.780 (4.311) & 19.736 (7.819) & 25.817 (10.679) & 28.634 (14.156) \\
    \quad Chronos-2 & \runnerup{11.590} (4.093) & \runnerup{18.309} (7.658) & \runnerup{24.649} (9.250) & \runnerup{27.809} (12.593) \\
    \quad \agentname{} & 12.291 (4.717) & 19.057 (8.742) & 25.245 (10.108) & 28.577 (12.989) \\
    \addlinespace
    \multicolumn{5}{l}{\textit{CGM only, full history}} \\
    \quad TimesFM-2.5 & \best{10.804} (3.713) & \runnerup{17.113} (7.298) & 23.109 (9.301) & 25.373 (12.535) \\
    \quad Moirai-2.0 & 11.179 (4.063) & 17.227 (7.563) & 23.259 (8.929) & 26.030 (12.192) \\
    \quad Timer$^{\dagger}$ & 11.891 (3.816) & 17.600 (7.207) & 23.240 (8.857) & 25.804 (12.342) \\
    \quad Chronos-2 & \runnerup{11.139} (4.038) & \best{17.027} (7.493) & \best{22.265} (8.475) & \runnerup{24.943} (11.496) \\
    \quad \agentname{} & 11.397 (4.320) & 17.408 (7.947) & \runnerup{22.628} (8.623) & \best{24.897} (11.267) \\
    \addlinespace
    \multicolumn{5}{l}{\textit{Multimodal, full history}} \\
    \quad Chronos-2 & \best{11.015} (4.140) & \best{16.629} (7.178) & \best{21.870} (8.168) & \best{24.221} (10.949) \\
    \quad \agentname{} & 11.418 (4.345) & 16.766 (8.001) & 21.978 (8.528) & 24.299 (11.358) \\
    \bottomrule
  \end{tabular}
  \vspace{2pt}

  {\footnotesize \textbf{Bold} marks the lowest RMSE and \underline{underlining} the second lowest within each input setting; only means are styled. $^{\dagger}$Timer preserves its upstream 96-step patch flooring (so 12 hours becomes 8 hours) and uses causal-mean imputation for missing historical CGM.}
\end{table}

\begin{table}[htbp]
  \caption{\textbf{CGMacros held-out forecasting in the post-meal situation.} The same 1,903 held-out origins as Table~\ref{tab:main-cgmacros-holdout}, restricted to the 300 origins labeled post-meal. Situation labels were attached after every forecast was scored; no forecaster saw them and no origin was selected or dropped on them. Within each participant, RMSE uses only that participant's origins in this situation; cells report the unweighted participant mean (sample standard deviation), in mg/dL.}
  \label{tab:cgmacros-holdout-stratum-post-meal}
  \centering
  \small
  \setlength{\tabcolsep}{4pt}
  \begin{tabular}{lrrrr}
    \toprule
    Method & 30 min & 60 min & 90 min & 120 min \\
    \midrule
    \multicolumn{5}{l}{\textit{Reference}} \\
    \quad Persistence & 30.784 (9.262) & 39.157 (18.489) & 43.479 (23.900) & 44.780 (27.697) \\
    \addlinespace
    \multicolumn{5}{l}{\textit{CGM only, 12 hours of history}} \\
    \quad TimesFM-2.5 & \best{26.346} (8.134) & \best{33.645} (15.404) & \best{34.887} (18.755) & \best{33.224} (19.600) \\
    \quad Moirai-2.0 & \runnerup{27.007} (9.072) & 35.251 (17.287) & 38.646 (22.768) & 38.503 (25.151) \\
    \quad Timer$^{\dagger}$ & 29.782 (11.482) & \runnerup{35.191} (17.008) & \runnerup{35.825} (20.696) & \runnerup{33.515} (20.388) \\
    \quad Chronos-2 & 27.414 (8.010) & 35.906 (17.173) & 39.460 (22.397) & 39.027 (25.542) \\
    \quad \agentname{} & 27.790 (8.381) & 36.223 (15.390) & 38.896 (17.704) & 36.742 (19.211) \\
    \addlinespace
    \multicolumn{5}{l}{\textit{CGM only, full history}} \\
    \quad TimesFM-2.5 & \runnerup{25.913} (7.882) & \runnerup{31.885} (13.758) & \best{31.102} (16.512) & 28.384 (16.516) \\
    \quad Moirai-2.0 & 26.221 (7.839) & 32.197 (13.629) & 31.977 (16.723) & 29.019 (16.442) \\
    \quad Timer$^{\dagger}$ & 26.443 (8.953) & 32.396 (14.657) & 31.962 (17.592) & 29.663 (17.034) \\
    \quad Chronos-2 & 26.315 (7.736) & 31.928 (13.812) & 31.478 (16.100) & \best{27.629} (15.108) \\
    \quad \agentname{} & \best{25.826} (8.154) & \best{31.025} (13.557) & \runnerup{31.227} (14.455) & \runnerup{27.643} (13.893) \\
    \addlinespace
    \multicolumn{5}{l}{\textit{Multimodal, full history}} \\
    \quad Chronos-2 & 25.848 (7.550) & 31.100 (13.504) & 30.371 (15.692) & \best{26.609} (14.590) \\
    \quad \agentname{} & \best{23.068} (4.761) & \best{26.472} (9.029) & \best{29.015} (10.657) & 27.008 (11.570) \\
    \bottomrule
  \end{tabular}
  \vspace{2pt}

  {\footnotesize \textbf{Bold} marks the lowest RMSE and \underline{underlining} the second lowest within each input setting; only means are styled. $^{\dagger}$Timer preserves its upstream 96-step patch flooring (so 12 hours becomes 8 hours) and uses causal-mean imputation for missing historical CGM.}
\end{table}

\begin{table}[htbp]
  \caption{\textbf{CGMacros held-out forecasting in the post-exercise situation.} The same 1,903 held-out origins as Table~\ref{tab:main-cgmacros-holdout}, restricted to the 325 origins labeled post-exercise. Situation labels were attached after every forecast was scored; no forecaster saw them and no origin was selected or dropped on them. Within each participant, RMSE uses only that participant's origins in this situation; cells report the unweighted participant mean (sample standard deviation), in mg/dL.}
  \label{tab:cgmacros-holdout-stratum-post-activity}
  \centering
  \small
  \setlength{\tabcolsep}{4pt}
  \begin{tabular}{lrrrr}
    \toprule
    Method & 30 min & 60 min & 90 min & 120 min \\
    \midrule
    \multicolumn{5}{l}{\textit{Reference}} \\
    \quad Persistence & 18.136 (8.220) & 28.654 (16.165) & 37.273 (25.090) & 40.046 (24.255) \\
    \addlinespace
    \multicolumn{5}{l}{\textit{CGM only, 12 hours of history}} \\
    \quad TimesFM-2.5 & \runnerup{15.359} (7.798) & 25.014 (12.495) & 32.365 (17.927) & 32.730 (19.018) \\
    \quad Moirai-2.0 & \best{14.995} (9.016) & \best{23.956} (14.580) & 31.975 (17.666) & 33.629 (20.251) \\
    \quad Timer$^{\dagger}$ & 15.677 (8.442) & \runnerup{24.071} (15.233) & \runnerup{30.717} (18.444) & \runnerup{31.687} (19.734) \\
    \quad Chronos-2 & 16.678 (7.138) & 26.577 (13.669) & 33.609 (20.606) & 34.354 (19.011) \\
    \quad \agentname{} & 16.383 (7.991) & 24.866 (12.252) & \best{30.711} (16.167) & \best{31.687} (18.076) \\
    \addlinespace
    \multicolumn{5}{l}{\textit{CGM only, full history}} \\
    \quad TimesFM-2.5 & \runnerup{15.070} (8.047) & 23.256 (13.770) & 28.926 (18.547) & 29.709 (19.646) \\
    \quad Moirai-2.0 & \best{14.535} (8.875) & \runnerup{22.514} (13.971) & \runnerup{28.815} (17.738) & 29.492 (18.927) \\
    \quad Timer$^{\dagger}$ & 15.134 (9.375) & 23.716 (13.857) & 29.279 (17.608) & 29.411 (18.559) \\
    \quad Chronos-2 & 15.739 (7.375) & 23.864 (12.738) & 29.144 (17.694) & \runnerup{29.367} (17.020) \\
    \quad \agentname{} & 15.215 (8.676) & \best{21.784} (12.827) & \best{26.302} (14.594) & \best{26.902} (16.894) \\
    \addlinespace
    \multicolumn{5}{l}{\textit{Multimodal, full history}} \\
    \quad Chronos-2 & \best{15.463} (7.734) & 22.486 (13.704) & 26.853 (16.649) & 27.145 (17.395) \\
    \quad \agentname{} & 15.646 (7.519) & \best{21.606} (11.819) & \best{25.625} (13.787) & \best{25.215} (14.675) \\
    \bottomrule
  \end{tabular}
  \vspace{2pt}

  {\footnotesize \textbf{Bold} marks the lowest RMSE and \underline{underlining} the second lowest within each input setting; only means are styled. $^{\dagger}$Timer preserves its upstream 96-step patch flooring (so 12 hours becomes 8 hours) and uses causal-mean imputation for missing historical CGM.}
\end{table}

\begin{table}[htbp]
  \caption{\textbf{CGMacros held-out forecasting in the meal + exercise situation.} The same 1,903 held-out origins as Table~\ref{tab:main-cgmacros-holdout}, restricted to the 361 origins labeled meal + exercise. Situation labels were attached after every forecast was scored; no forecaster saw them and no origin was selected or dropped on them. Within each participant, RMSE uses only that participant's origins in this situation; cells report the unweighted participant mean (sample standard deviation), in mg/dL.}
  \label{tab:cgmacros-holdout-stratum-compound}
  \centering
  \small
  \setlength{\tabcolsep}{4pt}
  \begin{tabular}{lrrrr}
    \toprule
    Method & 30 min & 60 min & 90 min & 120 min \\
    \midrule
    \multicolumn{5}{l}{\textit{Reference}} \\
    \quad Persistence & 26.867 (12.443) & 39.380 (18.906) & 40.101 (16.033) & 39.878 (16.410) \\
    \addlinespace
    \multicolumn{5}{l}{\textit{CGM only, 12 hours of history}} \\
    \quad TimesFM-2.5 & \runnerup{22.189} (7.766) & \best{33.313} (16.995) & \best{32.727} (17.060) & \runnerup{30.835} (18.338) \\
    \quad Moirai-2.0 & \best{21.248} (7.222) & \runnerup{34.472} (15.486) & \runnerup{34.440} (15.992) & 33.004 (17.916) \\
    \quad Timer$^{\dagger}$ & 26.043 (12.630) & 36.845 (21.400) & 34.868 (18.070) & \best{30.815} (16.197) \\
    \quad Chronos-2 & 23.304 (8.433) & 36.470 (16.714) & 35.852 (16.267) & 34.027 (19.812) \\
    \quad \agentname{} & 23.202 (8.302) & 36.609 (17.560) & 36.234 (16.201) & 35.006 (15.610) \\
    \addlinespace
    \multicolumn{5}{l}{\textit{CGM only, full history}} \\
    \quad TimesFM-2.5 & 20.984 (6.938) & 30.955 (15.382) & 29.066 (14.030) & 27.113 (14.466) \\
    \quad Moirai-2.0 & \runnerup{19.830} (6.429) & 30.360 (14.383) & 28.822 (13.315) & 27.008 (15.270) \\
    \quad Timer$^{\dagger}$ & 21.664 (8.462) & 31.856 (17.573) & 30.187 (15.437) & 28.865 (14.169) \\
    \quad Chronos-2 & \best{19.745} (5.797) & \best{29.338} (12.534) & \runnerup{28.654} (13.053) & \runnerup{26.747} (14.288) \\
    \quad \agentname{} & 20.010 (6.362) & \runnerup{30.219} (13.055) & \best{28.280} (12.165) & \best{26.498} (14.394) \\
    \addlinespace
    \multicolumn{5}{l}{\textit{Multimodal, full history}} \\
    \quad Chronos-2 & 21.016 (7.574) & 31.603 (17.107) & 30.025 (16.296) & \best{27.310} (15.172) \\
    \quad \agentname{} & \best{19.944} (5.447) & \best{29.047} (14.130) & \best{28.465} (13.174) & 27.462 (11.113) \\
    \bottomrule
  \end{tabular}
  \vspace{2pt}

  {\footnotesize \textbf{Bold} marks the lowest RMSE and \underline{underlining} the second lowest within each input setting; only means are styled. $^{\dagger}$Timer preserves its upstream 96-step patch flooring (so 12 hours becomes 8 hours) and uses causal-mean imputation for missing historical CGM.}
\end{table}

\subsection{The CGMacros result at each horizon}
\label{sec:additional-by-horizon}

\begin{figure}[htbp]
  \centering
  \includegraphics[width=\linewidth]{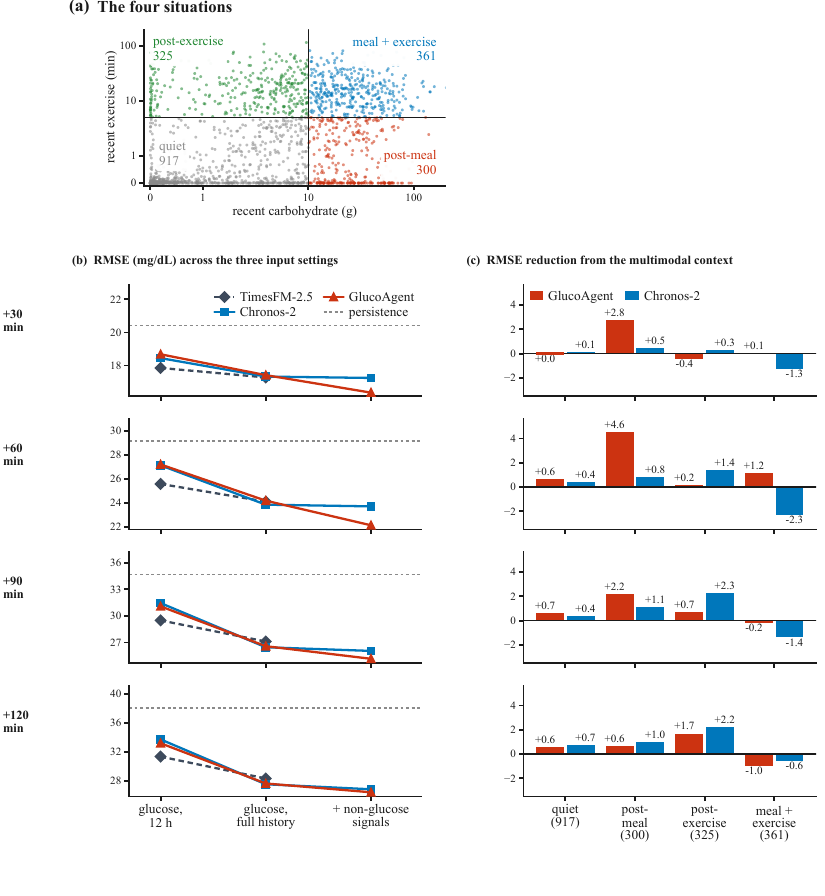}
  \caption{The CGMacros held-out results at each horizon, without averaging
  across horizons.  Panel (a) places every forecast origin by its recent
  carbohydrate and recent moderate exertion, each discounted by how long
  ago it happened, and the two fixed cuts that define quiet, post-meal,
  post-exercise and meal + exercise (for the scores and cuts see
  Appendix~\ref{sec:data-preparation-subsampling}); the labels were
  attached after every forecast was scored, and no method saw them.
  Panel (b) repeats the three input settings at each horizon.  Panel (c)
  reports the RMSE each method gains when the non-glucose signals are
  added, split by situation, on one shared scale.  The gain for
  \agentname{} is concentrated at 60 minutes: 2.1~mg/dL overall and
  4.6~mg/dL after a meal, against 1.1, 1.4 and 1.2~mg/dL overall at 30,
  90 and 120 minutes.  All panels use the full held-out set of 1,903
  forecast origins from nine participants.}
  \label{fig:by-horizon}
\end{figure}

The results figure in the main text averages the four horizons into one
number.  Figure~\ref{fig:by-horizon} reports the same quantities at each
horizon separately, and the average hides something real: the accuracy
\agentname{} gains from the non-glucose signals is concentrated at 60
minutes.

\clearpage

\section{What the Agent Does: Census and Remaining Case Studies}
\label{sec:behavior-appendix}

Nothing in the instruction tells \agentname{} which analysis to run, so
what it runs is an observation rather than a design choice, and this
appendix reports it in three ways: a census of the programs the agent
wrote, a census of the reasons it gave, and the case studies that
did not fit in the main text.

\subsection{Behavior by situation and stated reasons}
\label{sec:behavior-figure}

\begin{figure}[htbp]
  \centering
  \includegraphics[width=\linewidth]{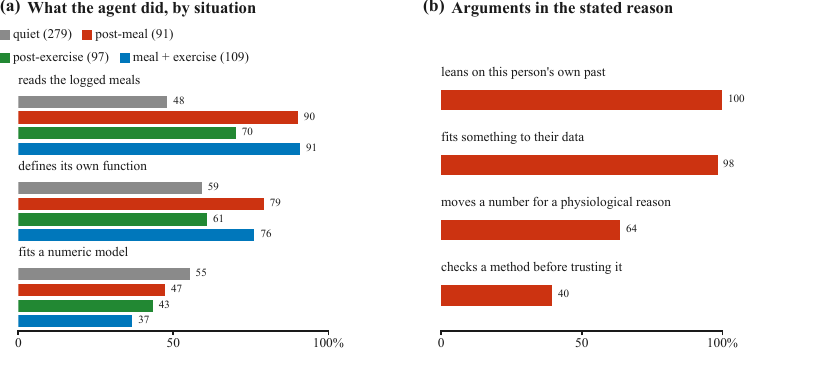}
  \caption{Behavior census on the audited subsample of 576 held-out
  CGMacros forecast origins from nine participants.  (a) The share of
  forecasts whose programs read the logged meals, defined a
  function of their own, or fitted a model, split by the four
  situations.  Reading the logged meals rises from 48\% on quiet origins
  to 90\% shortly after a meal, while fitting a model falls from
  55\% to 47\%.  (b) The share of forecasts whose submitted reason states
  each kind of argument.  These labels are assigned by a language-model
  judge reading the recorded reason against a fixed rubric, with every
  positive label carrying a verbatim span from the text.  Reliability was
  checked against a second judge from a different model family on 120 of
  the forecasts, and only rubric items the two judges agree on at least
  80\% of the time are shown; two of the six items fail that gate and are
  omitted.  A further 24 forecasts were re-read by hand, agreeing with
  the judge on 96.5\% of 144 judgments, but that audit was performed by
  the same assistant that built this analysis rather than by an
  independent rater, so it is a consistency check and not an independent
  validation.}
  \label{fig:behavior}
\end{figure}

Figure~\ref{fig:behavior} splits behavior two ways.  Panel (a) splits it
by situation, which the agent does not control, since the label is
computed from the person's context before the cutoff and no forecaster ever
receives it.  Reading the logged meals almost doubles between quiet origins
and origins shortly after a meal, and fitting a numeric model moves the
other way, so the agent falls back on model fitting when there is no recent
event to look up.  Panel (b) reports the arguments in its reasons.

\subsection{What the agent executed}
\label{sec:behavior-tables}

The tables below count what the agent actually ran, over the full
held-out set of 1,903 forecast origins rather than the audited subsample.
They are produced from the recorded run artifacts and from abstract
syntax trees of the submitted programs, so they count code rather than
claims.  Table~\ref{tab:behavior-program-inventory} states how many
programs were executed and how many parsed, which bounds every rate in
the tables that follow.  Table~\ref{tab:behavior-modality-evidence} asks
how strongly each signal is used, at four increasingly strict standards
of evidence.  Table~\ref{tab:behavior-stage-composition} reports when in
its own program sequence the agent does each thing.
Table~\ref{tab:behavior-software-stack} lists what it imported, and
Table~\ref{tab:behavior-estimators} lists every estimator and validation
symbol that defines the two behaviors we count, including the ones never
observed.

\begin{table}[htbp]
  \caption{What the agent executed across the full 1,903-origin CGMacros HOLDOUT cohort. Every quantity is counted from recorded run artifacts; the AST rows state the coverage within which the behavioural rates in the following tables hold.}
  \label{tab:behavior-program-inventory}
  \centering
  \small
  \begin{tabular}{lr}
    \toprule
    Quantity & Value \\
    \midrule
    Cases (participants) & 1903 (9) \\
    Programs requested / executed / successful & 14852 / 14846 / 14468 \\
    Executed failures and timeouts & 378 (166 timeouts) \\
    Failures followed by a later success & 373 / 378 \\
    Successful programs parsed for AST facts & 14468 / 14468 \\
    Successful programs per case, median [IQR] & 7 [6--9], range 4--14 \\
    Non-blank lines per program, median [IQR] & 11 [8--24], range 1--73 \\
    Cases authoring a function or class & 1227 / 1903 \\
    Subscripts with a literal key & 49.8\% \\
    Citation integrity (cited refs issued in-case) & 100.0\% \\
    \bottomrule
  \end{tabular}
  \vspace{2pt}

  {\footnotesize The literal-key fraction bounds the case-file field rates: accesses of the form \texttt{df[col]} with a variable key cannot be attributed to a modality, so every modality rate is a lower bound.}
\end{table}

\begin{table}[htbp]
  \caption{Use of each modality in the executed programs, cited programs, and submitted reasons, as a percentage of the 1,903 held-out forecast origins. \emph{Read} means a program subscripted a field of that modality; \emph{cited} means the submission cited such a program; \emph{named} means the submitted reason mentions the modality. The final column requires all three.}
  \label{tab:behavior-modality-evidence}
  \centering
  \small
  \begin{tabular}{lrrrr}
    \toprule
    Modality & Read & Cited & Named & Cited + named \\
    \midrule
    glucose & 100.0 & 100.0 & 99.6 & \textbf{99.6} \\
    heart rate & 96.3 & 95.2 & 55.8 & \textbf{54.2} \\
    activity & 96.4 & 95.5 & 76.7 & \textbf{73.8} \\
    meals & 67.1 & 65.5 & 99.5 & \textbf{65.1} \\
    baseline profile & 100.0 & 2.8 & 15.0 & \textbf{1.4} \\
    \bottomrule
  \end{tabular}
  \vspace{2pt}

  {\footnotesize The gap between \emph{read} and \emph{cited + named} is informative rather than noise. Heart rate is read in 96.3\% of cases but survives all three surfaces in 54.2\%. The baseline profile is cited in 2.8\% of cases and named in 15.0\%.}
\end{table}

\begin{table}[htbp]
  \caption{Share of the 1,903 HOLDOUT cases performing each operation in the early, middle and late third of \emph{their own} program sequence, so the table does not depend on how many programs a case ran. The final column is the median program number at which the operation first appears.}
  \label{tab:behavior-stage-composition}
  \centering
  \small
  \begin{tabular}{lrrrr}
    \toprule
    Operation & Early & Middle & Late & First \\
    \midrule
    reads glucose & 71.7 & 99.7 & 98.3 & 3 \\
    reads heart rate & 68.5 & 38.2 & 11.0 & 3 \\
    reads activity & 68.4 & 41.3 & 15.8 & 3 \\
    reads meals & 36.0 & 46.2 & 28.7 & 3 \\
    reads profile & 99.5 & 0.0 & 0.0 & 2 \\
    writes a function & 2.5 & 36.5 & 54.4 & 6 \\
    fits a model & 0.1 & 21.4 & 45.7 & 6 \\
    validates out of sample & 0.1 & 4.8 & 9.3 & 7 \\
    \bottomrule
  \end{tabular}
  \vspace{2pt}

  {\footnotesize The prompt prescribes no method, so the ordering is emergent. The baseline profile is read near the start and then dropped; glucose is revisited throughout; authoring, fitting and validation rise later in the sequence.}
\end{table}

\begin{table}[htbp]
  \caption{What the agent imported, as a percentage of the 1,903 HOLDOUT cases. Packages are shown first, then every module imported in at least four cases. Module names are taken verbatim from the import statement, so they are exact.}
  \label{tab:behavior-software-stack}
  \centering
  \small
  \begin{tabular}{lrrr}
    \toprule
    Package or module & Cases & Any use & Cited use \\
    \midrule
    \textbf{numpy} & & 100.0 & 100.0 \\
    \textbf{pandas} & & 91.3 & 90.9 \\
    \textbf{scikit-learn} & & 25.4 & 24.7 \\
    \textbf{scipy} & & 5.8 & 4.7 \\
    \textbf{statsmodels} & & 5.7 & 5.0 \\
    \addlinespace
    \texttt{numpy} & 1903 & 100.0 & 100.0 \\
    \texttt{toolbox.case\_data} & 1903 & 100.0 & 82.0 \\
    \texttt{toolbox.modalities} & 1903 & 100.0 & 11.5 \\
    \texttt{json} & 1899 & 99.8 & 74.0 \\
    \texttt{pandas} & 1737 & 91.3 & 90.9 \\
    \texttt{datetime} & 642 & 33.7 & 28.6 \\
    \texttt{sklearn.linear\_model} & 480 & 25.2 & 24.5 \\
    \texttt{sklearn.model\_selection} & 243 & 12.8 & 11.2 \\
    \texttt{numpy.linalg} & 102 & 5.4 & 4.4 \\
    \texttt{itertools} & 94 & 4.9 & 3.3 \\
    \texttt{scipy.optimize} & 92 & 4.8 & 3.8 \\
    \texttt{collections} & 82 & 4.3 & 3.0 \\
    \texttt{statsmodels.api} & 54 & 2.8 & 2.4 \\
    \texttt{sklearn.preprocessing} & 44 & 2.3 & 2.0 \\
    \texttt{sklearn.ensemble} & 35 & 1.8 & 1.7 \\
    \texttt{statsmodels.tsa.ar\_model} & 34 & 1.8 & 1.6 \\
    \texttt{sklearn.pipeline} & 32 & 1.7 & 1.5 \\
    \texttt{time} & 31 & 1.6 & 1.5 \\
    \texttt{sklearn.neighbors} & 22 & 1.2 & 1.1 \\
    \texttt{numpy.polynomial.polynomial} & 16 & 0.8 & 0.8 \\
    \texttt{pickle} & 16 & 0.8 & 0.5 \\
    \texttt{statsmodels.tsa.arima.model} & 16 & 0.8 & 0.8 \\
    \texttt{scipy.signal} & 15 & 0.8 & 0.7 \\
    \texttt{bisect} & 13 & 0.7 & 0.6 \\
    \texttt{math} & 11 & 0.6 & 0.5 \\
    \texttt{statsmodels.tsa.holtwinters} & 11 & 0.6 & 0.6 \\
    \texttt{sklearn.metrics} & 10 & 0.5 & 0.5 \\
    \texttt{numpy.polynomial} & 7 & 0.4 & 0.3 \\
    \texttt{warnings} & 6 & 0.3 & 0.3 \\
    \texttt{numpy.fft} & 5 & 0.3 & 0.1 \\
    \texttt{random} & 5 & 0.3 & 0.2 \\
    \texttt{scipy.stats} & 4 & 0.2 & 0.2 \\
    \texttt{statistics} & 4 & 0.2 & 0.1 \\
    \texttt{sys} & 4 & 0.2 & 0.1 \\
    \bottomrule
  \end{tabular}
  \vspace{2pt}

  {\footnotesize \emph{Cited use} restricts to programs the submission cited as evidence, so exploratory imports remain visible as a gap between the two columns.}
\end{table}

\begin{table}[htbp]
  \caption{Every symbol in the two printed sets that define \emph{fits a model} and \emph{validates out of sample}, with the number and percentage of the 1,903 HOLDOUT cases that called it. Membership of these sets is the whole definition of the two behaviors, so symbols never observed remain listed.}
  \label{tab:behavior-estimators}
  \centering
  \small
  \begin{tabular}{lrrr}
    \toprule
    Symbol & Cases & Any use & Cited use \\
    \midrule
    \multicolumn{4}{l}{\emph{Model fitting}} \\
    \quad \texttt{Ridge} & 464 & 24.4 & 23.5 \\
    \quad \texttt{polyfit} & 255 & 13.4 & 12.0 \\
    \quad \texttt{lstsq} & 223 & 11.7 & 10.6 \\
    \quad \texttt{curve\_fit} & 70 & 3.7 & 2.7 \\
    \quad \texttt{AutoReg} & 33 & 1.7 & 1.6 \\
    \quad \texttt{OLS} & 30 & 1.6 & 1.4 \\
    \quad \texttt{ARIMA} & 26 & 1.4 & 1.2 \\
    \quad \texttt{RidgeCV} & 19 & 1.0 & 0.8 \\
    \quad \texttt{KNeighborsRegressor} & 17 & 0.9 & 0.8 \\
    \quad \texttt{HistGradientBoostingRegressor} & 14 & 0.7 & 0.7 \\
    \quad \texttt{GradientBoostingRegressor} & 10 & 0.5 & 0.5 \\
    \quad \texttt{SARIMAX} & 10 & 0.5 & 0.4 \\
    \quad \texttt{ExponentialSmoothing} & 9 & 0.5 & 0.5 \\
    \quad \texttt{RandomForestRegressor} & 8 & 0.4 & 0.4 \\
    \quad \texttt{LinearRegression} & 3 & 0.2 & 0.1 \\
    \quad \texttt{Holt} & 2 & 0.1 & 0.1 \\
    \addlinespace
    \multicolumn{4}{l}{\emph{Out-of-sample validation}} \\
    \quad \texttt{KFold} & 143 & 7.5 & 6.1 \\
    \quad \texttt{TimeSeriesSplit} & 52 & 2.7 & 2.5 \\
    \quad \texttt{cross\_val\_predict} & 35 & 1.8 & 1.6 \\
    \quad \texttt{cross\_val\_score} & 35 & 1.8 & 1.6 \\
    \quad \texttt{GroupKFold} & 11 & 0.6 & 0.5 \\
    \quad \texttt{LeaveOneGroupOut} & 3 & 0.2 & 0.2 \\
    \quad \texttt{LeaveOneOut} & 1 & 0.1 & 0.1 \\
    \bottomrule
  \end{tabular}
  \vspace{2pt}

  {\footnotesize Defined but never called under \emph{model fitting}: \texttt{ElasticNet}, \texttt{Lasso}.  Defined but never called under \emph{out-of-sample validation}: \texttt{cross\_validate}, \texttt{train\_test\_split}. Naming a behavior only where a symbol uniquely implies it makes these rates auditable, but hand-written validation loops remain false negatives.}
\end{table}

\subsection{The remaining case studies}
\label{sec:behavior-cases}

\begin{figure}[htbp]
  \centering
  \includegraphics[width=\linewidth]{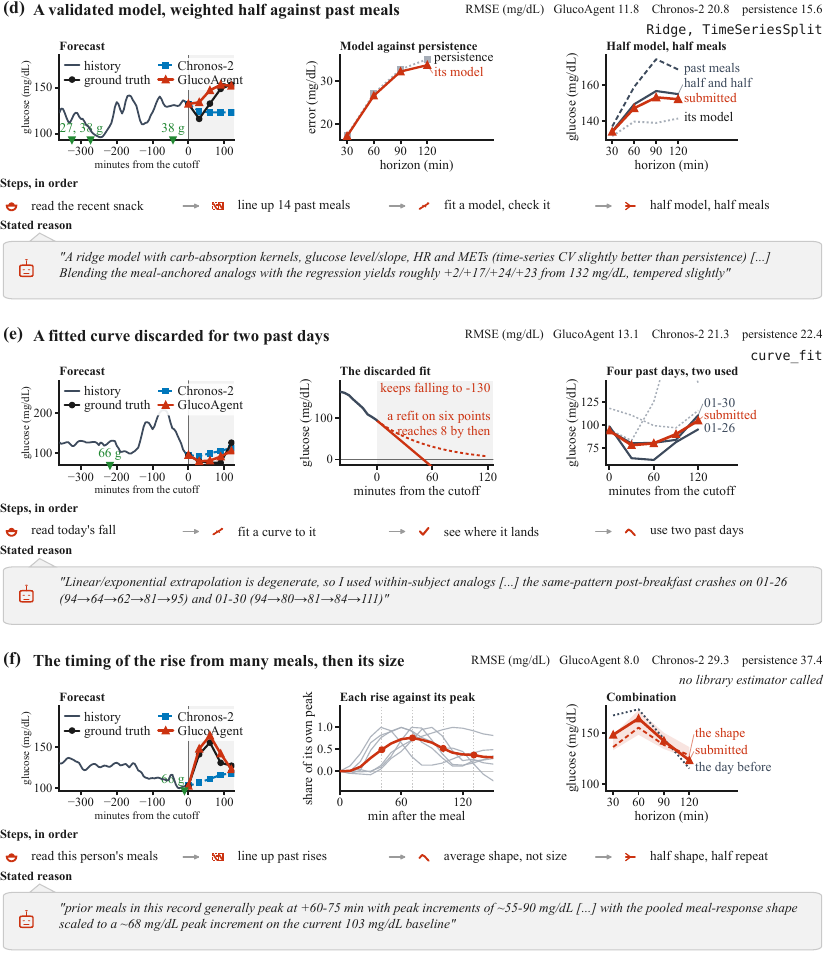}
  \caption{Case studies (d) to (f), the first half of the six forecasts
  that pass the recorded selection filter and are not shown in the main
  text.  Each row uses the same layout as the main-text figure: the
  person's context around the forecast cutoff, what the agent's own
  programs computed, in order, a verbatim fragment of the reason it
  submitted, and the forecast against the truth and against the
  multimodal foundation-model baseline.  The three rows show a ridge regression
  evaluated with time-series cross-validation and blended with past meal
  responses, a curve fit that is rejected in favor of two past days, and
  a forecast that works out when the rise arrives before working out how
  large it is.
  In (d) the fitted model runs only 0.2 to 1.4~mg/dL ahead of persistence
  at the four horizons; in (e) the two past days the agent did not use
  are drawn faint, and one of them rises to 212~mg/dL, above the chart;
  in (f) the middle chart keeps six of the nine rises the agent printed,
  selected by hand, with their average, and the shading on the right is
  the other peak sizes it tried.}
  \label{fig:cases-appendix-1}
\end{figure}

\begin{figure}[htbp]
  \centering
  \includegraphics[width=\linewidth]{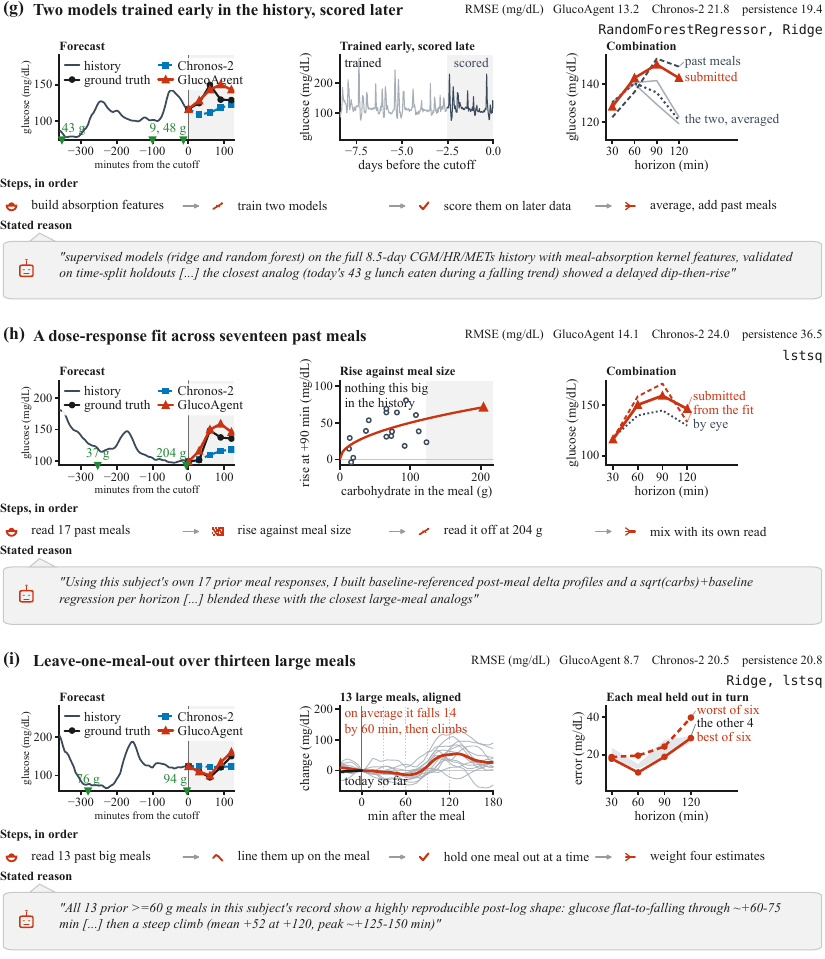}
  \caption{Case studies (g) to (i), the second half of the six forecasts
  that pass the recorded selection filter and are not shown in the main
  text.  The layout is the one used in
  Figure~\ref{fig:cases-appendix-1}.  The three rows are a model trained
  on the earlier history and scored on the later history, a fit of how
  this person's rise scales with the size of a meal, and a forecast built
  on the observation that this person dips for about an hour before
  rising.  In (g) the model is off by 12~mg/dL on the scored history and
  persistence by 19; in (h) the estimate labeled ``by eye'' is
  typed into the reason rather than computed; in (i) the right chart
  shows the best and worst of the six held-out meals, and the forecast
  used both.}
  \label{fig:cases-appendix-2}
\end{figure}

The case studies were selected by a filter recorded before any case was
read: the forecast had to beat the multimodal foundation-model baseline
by at least 5~mg/dL, have an RMSE of at most 15~mg/dL, and cover an
observed glucose swing of at least 25~mg/dL.  Nine of the 576 audited
origins clear all three conditions.  Three are shown in the main text.
Figures~\ref{fig:cases-appendix-1} and~\ref{fig:cases-appendix-2} show
the other six, so every case that passed the filter appears somewhere in
the paper and none was dropped after the fact.  The same filter reversed
selects seventeen origins where the agent loses to the baseline.  The
charts draw the agent's own recorded output, or, where a program printed
only a summary, the same computation re-run from the same context.

\clearpage

\section{The Established Benchmarks in Detail}
\label{sec:benchmarks-appendix}

This appendix supports Section~\ref{sec:results-benchmarks}: a check that
the fixed subsets rank the foundation models as the full populations
do, the OhioT1DM comparison under the CGM-LSM protocol, our reproduction of
the published baseline numbers, and why the two OhioT1DM RMSE values in the
literature are not comparable.

\subsection{The fixed subsets do not change the picture}
\label{sec:additional-subset-check}

We score the agent on fixed subsets of the glucose-only benchmarks to limit
the number of language-model calls
(Appendix~\ref{sec:data-preparation-subsampling}), which is defensible
only if the subset and the full population rank the methods the same way.
Table~\ref{tab:appendix-glucofm-subset-full-tsfm} runs the check on
GlucoFM-Bench, where every foundation model can be scored on both.  Over
the 36 cells the largest difference is 1.59~mg/dL, 28 are below
1~mg/dL and 20 are below 0.5~mg/dL, and the ranking holds: Timer is the
weakest model on every dataset in both populations, and the other three are
close enough that they exchange places by a few tenths on some datasets, in
both directions, while the subset never turns a clearly better model into a
clearly worse one.

\begin{table}[htbp]
  \caption{Sensitivity of reproduced GlucoFM-Bench TSFM results to the
  evaluation population. Full uses all eligible dense stride-1 origins;
  subset uses the evaluated selection of at most 32 origins per eligible
  participant. Cells report participant-level RMSE mean (sample standard
  deviation), in mg/dL, at the 30-minute horizon. $\Delta$ is subset minus
  full.}
  \label{tab:appendix-glucofm-subset-full-tsfm}
  \centering
  \scriptsize
  \setlength{\tabcolsep}{3.0pt}
  \resizebox{\textwidth}{!}{%
  \begin{tabular}{llrrrcc}
    \toprule
    Dataset & Model & Full dense & Subset $K=32$ & $\Delta$ & Full $n$ & Subset $n$ \\
    \midrule
    \multirow{4}{*}{Hall2018} & Chronos-2 & 13.41 (6.93) & 13.27 (6.72) & -0.15 & 47 & 47 \\
     & Moirai-2.0 & 13.16 (6.06) & 12.96 (6.94) & -0.20 & 48 & 47 \\
     & Timer & 14.05 (7.11) & 14.27 (8.10) & +0.22 & 48 & 47 \\
     & TimesFM-2.5 & 12.96 (5.83) & 12.92 (6.49) & -0.04 & 48 & 47 \\
    \addlinespace
    \multirow{4}{*}{BIG IDEAS Lab} & Chronos-2 & 15.17 (3.04) & 14.21 (3.84) & -0.96 & 16 & 16 \\
     & Moirai-2.0 & 15.18 (3.24) & 14.29 (3.97) & -0.89 & 16 & 16 \\
     & Timer & 16.24 (4.78) & 15.95 (5.19) & -0.29 & 16 & 16 \\
     & TimesFM-2.5 & 14.88 (3.00) & 13.78 (3.80) & -1.10 & 16 & 16 \\
    \addlinespace
    \multirow{4}{*}{ShanghaiT1DM} & Chronos-2 & 12.82 (3.41) & 12.51 (5.27) & -0.30 & 12 & 12 \\
     & Moirai-2.0 & 12.50 (3.18) & 11.89 (4.56) & -0.62 & 12 & 12 \\
     & Timer & 19.10 (6.16) & 19.59 (9.00) & +0.49 & 12 & 12 \\
     & TimesFM-2.5 & 12.11 (3.45) & 12.06 (5.22) & -0.05 & 12 & 12 \\
    \addlinespace
    \multirow{4}{*}{ShanghaiT2DM} & Chronos-2 & 11.94 (3.35) & 11.77 (3.80) & -0.17 & 99 & 99 \\
     & Moirai-2.0 & 11.96 (3.45) & 11.47 (3.51) & -0.48 & 99 & 99 \\
     & Timer & 15.60 (4.94) & 15.40 (5.76) & -0.20 & 99 & 99 \\
     & TimesFM-2.5 & 11.47 (3.31) & 11.14 (3.66) & -0.32 & 99 & 99 \\
    \addlinespace
    \multirow{4}{*}{HUPA-UCM} & Chronos-2 & 18.07 (5.49) & 18.04 (7.57) & -0.03 & 25 & 25 \\
     & Moirai-2.0 & 18.21 (5.45) & 17.12 (6.86) & -1.09 & 25 & 25 \\
     & Timer & 24.59 (6.58) & 24.49 (7.81) & -0.10 & 25 & 25 \\
     & TimesFM-2.5 & 17.67 (5.19) & 17.35 (7.42) & -0.32 & 25 & 25 \\
    \addlinespace
    \multirow{4}{*}{T1DM-UOM} & Chronos-2 & 22.24 (4.92) & 22.01 (5.53) & -0.22 & 17 & 17 \\
     & Moirai-2.0 & 22.55 (4.91) & 22.62 (6.08) & +0.06 & 17 & 17 \\
     & Timer & 25.83 (5.33) & 24.99 (6.47) & -0.84 & 17 & 17 \\
     & TimesFM-2.5 & 21.89 (4.83) & 22.34 (5.83) & +0.45 & 17 & 17 \\
    \addlinespace
    \multirow{4}{*}{Bris-T1D Open} & Chronos-2 & 19.42 (5.62) & 18.31 (8.33) & -1.11 & 20 & 20 \\
     & Moirai-2.0 & 19.64 (5.81) & 18.05 (7.94) & -1.59 & 20 & 20 \\
     & Timer & 22.81 (6.17) & 21.35 (10.26) & -1.45 & 20 & 20 \\
     & TimesFM-2.5 & 19.00 (5.58) & 17.89 (8.44) & -1.11 & 20 & 20 \\
    \addlinespace
    \multirow{4}{*}{AZT1D} & Chronos-2 & 20.77 (3.34) & 20.52 (5.61) & -0.25 & 25 & 25 \\
     & Moirai-2.0 & 21.23 (3.34) & 20.30 (5.61) & -0.93 & 25 & 25 \\
     & Timer & 23.83 (3.93) & 23.17 (5.42) & -0.66 & 25 & 25 \\
     & TimesFM-2.5 & 20.53 (3.29) & 19.69 (5.15) & -0.84 & 25 & 25 \\
    \addlinespace
    \multirow{4}{*}{OhioT1DM} & Chronos-2 & 19.72 (2.81) & 20.77 (4.90) & +1.05 & 12 & 12 \\
     & Moirai-2.0 & 20.24 (2.46) & 20.28 (4.65) & +0.04 & 12 & 12 \\
     & Timer & 24.58 (2.89) & 25.46 (5.41) & +0.88 & 12 & 12 \\
     & TimesFM-2.5 & 19.62 (2.66) & 20.88 (4.50) & +1.26 & 12 & 12 \\
    \bottomrule
  \end{tabular}}
\end{table}

\subsection{OhioT1DM under the CGM-LSM protocol}
\label{sec:additional-cgmlsm}

Table~\ref{tab:appendix-cgmlsm} gives the OhioT1DM comparison that
Section~\ref{sec:results-benchmarks} summarizes, scored with the CGM-LSM
evaluator on the matched 1{,}536-instance subset; the published CGM-LSM row
is transcribed from that paper (for why its numbers are not comparable
with the GlucoFM-Bench OhioT1DM column see Appendix~\ref{sec:ohio-metric-gap}).

\begin{table}[htbp]
  \caption{OhioT1DM under the CGM-LSM-matched protocol. Exact local rows use
  the evaluated subset ($n=1536$). Cells report mean
  per-instance cumulative-window rMSE (Student-$t$ 95\% confidence half-width),
  in mg/dL; the parenthesized quantity is not a standard deviation.}
  \label{tab:appendix-cgmlsm}
  \centering
  \small
  \setlength{\tabcolsep}{6pt}
  \begin{tabular}{lccc}
    \toprule
    Method & 0--30 min & 0--60 min & 0--120 min \\
    \midrule
    CGM-LSM$^{\ast}$ (reported)
      & 9.024 (0.168) & 15.895 (0.283) & 26.876 (0.430) \\
    Chronos-2 (zero-shot)
      & \runnerup{9.059} (0.398) & \runnerup{15.865} (0.676) & \runnerup{25.780} (0.982) \\
    Moirai-2.0 (zero-shot)
      & 9.232 (0.417) & 16.218 (0.701) & 26.726 (1.035) \\
    Timer$^{\dagger}$ (zero-shot)
      & 10.485 (0.426) & 17.204 (0.682) & 26.825 (0.930) \\
    TimesFM-2.5 (zero-shot)
      & \best{8.902} (0.397) & \best{15.686} (0.667) & \best{25.488} (0.967) \\
    Persistence
      & 11.089 (0.490) & 18.649 (0.771) & 29.537 (1.113) \\
    \midrule
    \agentname{}
      & 9.177 (0.409) & 16.030 (0.685) & 25.835 (1.001) \\
    \bottomrule
  \end{tabular}
  \vspace{2pt}

  {\footnotesize $^{\ast}$The CGM-LSM row is the sole copied numerical row
  and is transcribed from the publication's results notebook
  (\url{https://github.com/JHU-CDHAI/cgmlsm}); the model itself was not run
  here. It is not evaluated on the 1536-instance subset. Among the exactly matched local
  rows, \textbf{bold} is best and \underline{underlining} is second best;
  only means are styled. $^{\dagger}$Timer uses the
  last 288 of 289 context points because its standard input patch is 96 steps.}
\end{table}

\subsection{Reproducing Published Results}
\label{sec:reproducing-published-results}

Before comparing against a published number we reproduce it, so that a
difference in the main tables is a difference in forecasting rather than
in setup; everything in this subsection checks the baselines, and
\agentname{} has no row here because it has no published score to
reproduce and is scored on the fixed subsets of
Appendix~\ref{sec:data-preparation-subsampling}.

\subsubsection{Published-setting reproduction fidelity}

\begingroup
\scriptsize
\setlength{\tabcolsep}{3.5pt}
\begin{longtable}{llrrrr}
  \caption{Reproduction of accessible per-dataset zero-shot cells for the
  datasets retained in this paper from GlucoFM-Bench Table 2. RMSE is in
  mg/dL at 12 hours of context and a
  30-minute horizon; values are participant-level mean (sample standard
  deviation).}
  \label{tab:appendix-glucofm-fidelity}\\
  \toprule
  Dataset & Model & Published & Reproduced & $\Delta$ & $n$ \\
  \midrule
  \endfirsthead

  \multicolumn{6}{l}{\footnotesize\emph{Table
  \ref{tab:appendix-glucofm-fidelity} continued}}\\
  \toprule
  Dataset & Model & Published & Reproduced & $\Delta$ & $n$ \\
  \midrule
  \endhead

  \midrule
  \multicolumn{6}{r}{\footnotesize\emph{Continued on next page}}\\
  \endfoot

  \bottomrule
  \endlastfoot

  \multirow{5}{*}{Hall2018}
    & Persistence & 13.65 (4.52) & 13.66 (4.52) & +0.006 & 52 \\
    & Chronos-2 & 13.24 (6.96) & 13.41 (6.93) & +0.173 & 47 \\
    & Moirai-2.0 & 13.16 (6.06) & 13.16 (6.06) & -0.004 & 48 \\
    & Timer & 14.05 (7.11) & 14.05 (7.11) & +0.002 & 48 \\
    & TimesFM-2.5 & 12.96 (5.83) & 12.96 (5.83) & -0.001 & 48 \\
  \addlinespace
  \multirow{5}{*}{BIG IDEAS Lab}
    & Persistence & 16.91 (4.03) & 16.92 (4.03) & +0.006 & 16 \\
    & Chronos-2 & 15.17 (3.04) & 15.17 (3.04) & -0.001 & 16 \\
    & Moirai-2.0 & 15.18 (3.24) & 15.18 (3.24) & +0.005 & 16 \\
    & Timer & 16.24 (4.78) & 16.24 (4.78) & -0.001 & 16 \\
    & TimesFM-2.5 & 14.88 (3.00) & 14.88 (3.00) & -0.003 & 16 \\
  \addlinespace
  \multirow{5}{*}{ShanghaiT1DM}
    & Persistence & 17.18 (6.57) & 17.19 (6.57) & +0.009 & 12 \\
    & Chronos-2 & 12.82 (3.41) & 12.82 (3.41) & -0.005 & 12 \\
    & Moirai-2.0 & 12.50 (3.18) & 12.50 (3.18) & +0.004 & 12 \\
    & Timer & 19.10 (6.16) & 19.10 (6.16) & -0.003 & 12 \\
    & TimesFM-2.5 & 12.11 (3.45) & 12.11 (3.45) & +0.004 & 12 \\
  \addlinespace
  \multirow{5}{*}{ShanghaiT2DM}
    & Persistence & 15.03 (4.14) & 15.04 (4.14) & +0.007 & 100 \\
    & Chronos-2 & 11.94 (3.35) & 11.94 (3.35) & +0.001 & 99 \\
    & Moirai-2.0 & 11.96 (3.45) & 11.96 (3.45) & -0.004 & 99 \\
    & Timer & 15.60 (4.94) & 15.60 (4.94) & +0.001 & 99 \\
    & TimesFM-2.5 & 11.47 (3.31) & 11.47 (3.31) & -0.003 & 99 \\
  \addlinespace
  \multirow{5}{*}{HUPA-UCM}
    & Persistence & 22.27 (6.15) & 22.27 (6.15) & +0.004 & 25 \\
    & Chronos-2 & 18.07 (5.49) & 18.07 (5.49) & +0.003 & 25 \\
    & Moirai-2.0 & 18.21 (5.45) & 18.21 (5.45) & -0.002 & 25 \\
    & Timer & 24.59 (6.58) & 24.59 (6.58) & +0.000 & 25 \\
    & TimesFM-2.5 & 17.67 (5.19) & 17.67 (5.19) & +0.004 & 25 \\
  \addlinespace
  \multirow{5}{*}{CGMacros}
    & Persistence & 19.44 (4.88) & 19.44 (4.88) & -0.002 & 45 \\
    & Chronos-2 & 16.85 (4.16) & 16.85 (4.16) & -0.000 & 45 \\
    & Moirai-2.0 & 16.81 (4.36) & 16.81 (4.36) & -0.004 & 45 \\
    & Timer & 17.42 (4.88) & 17.42 (4.88) & -0.004 & 45 \\
    & TimesFM-2.5 & 16.28 (4.18) & 16.28 (4.18) & -0.002 & 45 \\
  \addlinespace
  \multirow{5}{*}{T1DM-UOM}
    & Persistence & 24.95 (5.17) & 24.95 (5.17) & +0.003 & 17 \\
    & Chronos-2 & 22.23 (4.92) & 22.24 (4.92) & +0.005 & 17 \\
    & Moirai-2.0 & 22.55 (4.91) & 22.55 (4.91) & +0.004 & 17 \\
    & Timer & 25.83 (5.33) & 25.83 (5.33) & +0.004 & 17 \\
    & TimesFM-2.5 & 21.89 (4.83) & 21.89 (4.83) & -0.002 & 17 \\
  \addlinespace
  \multirow{5}{*}{Bris-T1D Open}
    & Persistence & 22.24 (6.47) & 22.24 (6.47) & +0.002 & 20 \\
    & Chronos-2 & 19.42 (5.62) & 19.42 (5.62) & +0.003 & 20 \\
    & Moirai-2.0 & 19.64 (5.81) & 19.64 (5.81) & -0.004 & 20 \\
    & Timer & 22.81 (6.17) & 22.81 (6.17) & -0.004 & 20 \\
    & TimesFM-2.5 & 19.00 (5.58) & 19.00 (5.58) & +0.002 & 20 \\
  \addlinespace
  \multirow{5}{*}{AZT1D}
    & Persistence & 23.36 (3.39) & 23.36 (3.39) & -0.002 & 25 \\
    & Chronos-2 & 20.77 (3.34) & 20.77 (3.34) & +0.004 & 25 \\
    & Moirai-2.0 & 21.23 (3.34) & 21.23 (3.34) & +0.002 & 25 \\
    & Timer & 23.83 (3.93) & 23.83 (3.93) & +0.003 & 25 \\
    & TimesFM-2.5 & 20.53 (3.29) & 20.53 (3.29) & -0.005 & 25 \\
  \addlinespace
  \multirow{5}{*}{OhioT1DM}
    & Persistence & 23.42 (2.97) & 23.42 (2.97) & -0.000 & 12 \\
    & Chronos-2 & 19.72 (2.81) & 19.72 (2.81) & -0.003 & 12 \\
    & Moirai-2.0 & 20.24 (2.46) & 20.24 (2.46) & -0.000 & 12 \\
    & Timer & 24.58 (2.89) & 24.58 (2.89) & +0.003 & 12 \\
    & TimesFM-2.5 & 19.63 (2.84) & 19.62 (2.66) & -0.008 & 12 \\
\end{longtable}
\endgroup

The reproduction covers all 50 published cells that admit evaluation for the
datasets retained in this paper. The median absolute deviation is 0.003 mg/dL,
the mean is 0.007 mg/dL, and the maximum is 0.173 mg/dL. The
Hall2018/Chronos-2 residual follows the upstream
47-participant sequence-splitting path. OhioT1DM/TimesFM-2.5 matches in mean but
not in participant spread; this is the only unexplained residual.

\subsubsection{Full multi-horizon GlucoFM-Bench reproduction}

\begingroup
\scriptsize
\setlength{\tabcolsep}{0.8pt}
\begin{longtable}{llrrrrr}
  \caption{Full zero-shot GlucoFM-Bench reproduction with the available
  published comparison. Entries are participant-level RMSE mean (sample
  standard deviation), in mg/dL, with a 12-hour context and dense stride-1
  origins. Reproduced horizon columns score the final step; the final column is
  the only per-dataset horizon published by GlucoFM-Bench (Table 2).}
  \label{tab:appendix-glucofm-full}\\
  \toprule
  Dataset & Method & \multicolumn{4}{c}{Reproduced} & Published \\
  \cmidrule(lr){3-6}\cmidrule(l){7-7}
  & & 15 min & 30 min & 60 min & 90 min & 30 min \\
  \midrule
  \endfirsthead

  \multicolumn{7}{l}{\footnotesize\emph{Table
  \ref{tab:appendix-glucofm-full} continued}}\\
  \toprule
  Dataset & Method & \multicolumn{4}{c}{Reproduced} & Published \\
  \cmidrule(lr){3-6}\cmidrule(l){7-7}
  & & 15 min & 30 min & 60 min & 90 min & 30 min \\
  \midrule
  \endhead

  \midrule
  \multicolumn{7}{r}{\footnotesize\emph{Continued on next page}}\\
  \endfoot

  \bottomrule
  \endlastfoot

  \multirow{5}{*}{AZT1D}
    & Persistence & 14.11 (2.20) & 23.36 (3.39) & 36.94 (5.55) & 46.09 (7.54) & 23.36 (3.39) \\
    & Chronos-2 & 12.04 (2.42) & 20.77 (3.34) & 34.14 (5.29) & 42.60 (7.28) & 20.77 (3.34) \\
    & TimesFM-2.5 & 11.87 (2.47) & 20.53 (3.29) & 33.45 (5.22) & 41.66 (7.32) & 20.53 (3.29) \\
    & Timer & 14.23 (2.52) & 23.83 (3.93) & 36.68 (6.75) & 43.71 (9.25) & 23.83 (3.93) \\
    & Moirai-2.0 & 12.25 (2.44) & 21.23 (3.34) & 35.09 (5.26) & 44.27 (7.48) & 21.23 (3.34) \\
  \addlinespace
  \multirow{5}{*}{BIG IDEAS Lab}
    & Persistence & 10.28 (2.16) & 16.92 (4.03) & 23.14 (7.02) & 25.49 (9.07) & 16.91 (4.03) \\
    & Chronos-2 & 8.05 (1.49) & 15.17 (3.04) & 20.81 (5.33) & 22.38 (6.82) & 15.17 (3.04) \\
    & TimesFM-2.5 & 7.72 (1.38) & 14.88 (3.00) & 20.54 (5.01) & 21.84 (6.30) & 14.88 (3.00) \\
    & Timer & 10.01 (2.96) & 16.24 (4.78) & 20.65 (6.28) & 21.71 (6.99) & 16.24 (4.78) \\
    & Moirai-2.0 & 8.04 (1.44) & 15.18 (3.24) & 21.05 (5.77) & 22.66 (6.73) & 15.18 (3.24) \\
  \addlinespace
  \multirow{5}{*}{Bris-T1D Open}
    & Persistence & 12.82 (3.78) & 22.24 (6.47) & 35.92 (10.47) & 44.89 (13.13) & 22.24 (6.47) \\
    & Chronos-2 & 10.42 (3.32) & 19.42 (5.62) & 33.47 (9.33) & 42.47 (11.88) & 19.42 (5.62) \\
    & TimesFM-2.5 & 10.09 (3.39) & 19.00 (5.58) & 32.66 (9.13) & 41.46 (11.66) & 19.00 (5.58) \\
    & Timer & 13.01 (3.55) & 22.81 (6.17) & 35.92 (9.88) & 43.25 (12.11) & 22.81 (6.17) \\
    & Moirai-2.0 & 10.52 (3.47) & 19.64 (5.81) & 33.89 (9.64) & 43.01 (12.22) & 19.64 (5.81) \\
  \addlinespace
  \multirow{5}{*}{CGMacros$^{\dagger}$}
    & Persistence & 11.93 (2.73) & 19.44 (4.88) & 27.60 (9.12) & 32.67 (12.65) & 19.44 (4.88) \\
    & Chronos-2 & 9.72 (2.30) & 16.85 (4.16) & 23.43 (7.32) & 26.27 (9.81) & 16.85 (4.16) \\
    & TimesFM-2.5 & 9.39 (2.29) & 16.28 (4.18) & 22.40 (7.21) & 25.02 (9.41) & 16.28 (4.18) \\
    & Timer & 10.80 (2.74) & 17.42 (4.88) & 23.20 (8.25) & 25.84 (10.36) & 17.42 (4.88) \\
    & Moirai-2.0 & 9.69 (2.37) & 16.81 (4.36) & 23.46 (7.67) & 26.55 (10.14) & 16.81 (4.36) \\
  \addlinespace
  \multirow{5}{*}{HUPA-UCM}
    & Persistence & 12.33 (3.29) & 22.27 (6.15) & 37.66 (10.69) & 48.77 (13.75) & 22.27 (6.15) \\
    & Chronos-2 & 8.87 (2.87) & 18.07 (5.49) & 34.41 (10.84) & 46.40 (14.87) & 18.07 (5.49) \\
    & TimesFM-2.5 & 8.45 (2.96) & 17.67 (5.19) & 34.38 (10.24) & 46.86 (14.71) & 17.67 (5.19) \\
    & Timer & 13.69 (3.69) & 24.59 (6.58) & 41.18 (11.39) & 52.03 (14.95) & 24.59 (6.58) \\
    & Moirai-2.0 & 8.95 (2.96) & 18.21 (5.45) & 35.44 (10.78) & 48.41 (14.99) & 18.21 (5.45) \\
  \addlinespace
  \multirow{5}{*}{Hall2018}
    & Persistence & 8.65 (2.60) & 13.66 (4.52) & 18.77 (7.74) & 21.46 (10.01) & 13.65 (4.52) \\
    & Chronos-2 & 8.14 (3.71) & 13.41 (6.93) & 18.90 (11.31) & 21.57 (12.77) & 13.24 (6.96) \\
    & TimesFM-2.5 & 7.96 (3.13) & 12.96 (5.83) & 17.99 (9.46) & 20.56 (11.55) & 12.96 (5.83) \\
    & Timer & 9.09 (4.21) & 14.05 (7.11) & 18.66 (10.49) & 20.87 (12.55) & 14.05 (7.11) \\
    & Moirai-2.0 & 8.10 (3.33) & 13.16 (6.06) & 18.51 (10.19) & 21.03 (12.49) & 13.16 (6.06) \\
  \addlinespace
  \multirow{5}{*}{OhioT1DM}
    & Persistence & 13.59 (1.67) & 23.42 (2.97) & 38.58 (4.87) & 49.42 (5.98) & 23.42 (2.97) \\
    & Chronos-2 & 10.89 (2.03) & 19.72 (2.81) & 34.33 (4.24) & 44.78 (5.23) & 19.72 (2.81) \\
    & TimesFM-2.5 & 10.81 (1.99) & 19.62 (2.66) & 34.03 (4.27) & 44.13 (5.39) & 19.63 (2.84) \\
    & Timer & 14.17 (2.04) & 24.58 (2.89) & 39.76 (4.21) & 49.25 (5.31) & 24.58 (2.89) \\
    & Moirai-2.0 & 11.18 (1.89) & 20.24 (2.46) & 35.57 (3.84) & 46.61 (4.79) & 20.24 (2.46) \\
  \addlinespace
  \multirow{5}{*}{ShanghaiT1DM}
    & Persistence & 9.11 (3.32) & 17.19 (6.57) & 29.96 (12.58) & 39.83 (17.71) & 17.18 (6.57) \\
    & Chronos-2 & 5.42 (1.37) & 12.82 (3.41) & 26.55 (8.50) & 37.82 (13.91) & 12.82 (3.41) \\
    & TimesFM-2.5 & 4.76 (1.19) & 12.11 (3.45) & 26.49 (9.07) & 37.99 (14.50) & 12.11 (3.45) \\
    & Timer & 10.30 (3.71) & 19.10 (6.16) & 33.65 (11.33) & 44.21 (15.68) & 19.10 (6.16) \\
    & Moirai-2.0 & 5.27 (1.29) & 12.50 (3.18) & 27.14 (8.29) & 39.36 (13.86) & 12.50 (3.18) \\
  \addlinespace
  \multirow{5}{*}{ShanghaiT2DM}
    & Persistence & 8.11 (2.16) & 15.04 (4.14) & 25.12 (7.45) & 32.10 (9.99) & 15.03 (4.14) \\
    & Chronos-2 & 5.24 (1.46) & 11.94 (3.35) & 22.72 (7.00) & 29.66 (10.25) & 11.94 (3.35) \\
    & TimesFM-2.5 & 4.72 (1.41) & 11.47 (3.31) & 22.59 (7.16) & 29.80 (10.32) & 11.47 (3.31) \\
    & Timer & 8.30 (2.91) & 15.60 (4.94) & 25.93 (8.68) & 31.70 (11.51) & 15.60 (4.94) \\
    & Moirai-2.0 & 5.11 (1.52) & 11.96 (3.45) & 23.49 (7.13) & 31.03 (10.30) & 11.96 (3.45) \\
  \addlinespace
  \multirow{5}{*}{T1DM-UOM}
    & Persistence & 14.37 (2.88) & 24.95 (5.17) & 39.56 (8.13) & 48.92 (10.02) & 24.95 (5.17) \\
    & Chronos-2 & 11.72 (2.54) & 22.24 (4.92) & 37.43 (7.81) & 46.39 (9.77) & 22.23 (4.92) \\
    & TimesFM-2.5 & 11.51 (2.49) & 21.89 (4.83) & 36.62 (7.54) & 45.33 (9.38) & 21.89 (4.83) \\
    & Timer & 14.48 (3.23) & 25.83 (5.33) & 40.44 (8.29) & 48.42 (10.28) & 25.83 (5.33) \\
    & Moirai-2.0 & 11.89 (2.51) & 22.55 (4.91) & 38.33 (7.92) & 48.07 (10.07) & 22.55 (4.91) \\
\end{longtable}
\endgroup

$^{\dagger}$The full CGMacros row uses the external all-45 cohort and is not a
held-out evaluation set.
GlucoFM-Bench does not publish per-dataset 15-, 60-, or 90-minute cells. Its
across-horizon tables instead pool 15 datasets. Three of them, UCHTT1DM,
DiaTrend, and T1DEXI, are unavailable here, so those aggregate rows are not a
like-for-like comparison and are omitted.

\subsubsection{Exact CGM-LSM-matched comparison}

\begin{table}[htbp]
  \caption{TSFM comparison with CGM-LSM on its exact OhioT1DM evaluation
  protocol. The zero-shot and persistence rows are reproduced on the same
  9,214-instance set. Cells report mean rMSE (Student-$t$ 95\% confidence
  half-width), in mg/dL.}
  \label{tab:appendix-cgmlsm-exact}
  \centering
  \footnotesize
  \setlength{\tabcolsep}{3pt}
  \begin{tabular}{llccc}
    \toprule
    Method & Result source & 0--30 min & 0--60 min & 0--120 min \\
    \midrule
    CGM-LSM$^{\ast}$ & reported, trained
      & 9.024 (0.168) & 15.895 (0.283) & 26.876 (0.43) \\
    Chronos-2 & reproduced, zero-shot
      & 9.007 (0.164) & 15.772 (0.275) & 25.654 (0.397) \\
    TimesFM-2.5 & reproduced, zero-shot
      & 8.864 (0.165) & 15.622 (0.273) & 25.460 (0.393) \\
    Persistence & reproduced
      & 11.198 (0.199) & 18.619 (0.315) & 29.447 (0.453) \\
    \bottomrule
  \end{tabular}
  \vspace{2pt}

  {\footnotesize $^{\ast}$CGM-LSM values, including the parenthesized
  uncertainty, are transcribed from the paper's Table~2; CGM-LSM itself was
  not run here. Its code and results notebook are at
  \url{https://github.com/JHU-CDHAI/cgmlsm}. Although the table caption
  says confidence-interval ``width,'' the released notebook computes the
  Student-$t$ half-width $t_{0.975,n-1}s/\sqrt{n}$. All reproduced rows use the
  same computation; see Section~\ref{sec:ohio-metric-gap}.
  \Agentname{} is scored on the matched 1,536-instance subset reported in
  Table~\ref{tab:appendix-cgmlsm} rather than on this full 9,214-instance
  population, so it has no row here.}
\end{table}

The CGM-LSM origin filter reads both the 289-step history and 24-step target
window. Its target-aware exclusions remove 73 of 9,287 otherwise eligible
spans. This is retained only to reproduce the published experimental condition;
post-cutoff values are not exposed to any forecasting method.

\subsection{Why the OhioT1DM RMSE Values Are Not Directly Comparable}
\label{sec:ohio-metric-gap}

GlucoFM-Bench and CGM-LSM both report an OhioT1DM RMSE at 30 minutes, and
the CGM-LSM value is about half the size, but the two names cover different
quantities computed on different origins, so the gap says nothing about
forecasting accuracy.  Let $e_{sij}$ be the error for participant $s$,
forecast origin $i$ and five-minute step $j$.  GlucoFM-Bench scores only
the endpoint $j=6$, pools the squared errors within a participant and
weights participants equally:
\begin{equation}
  R_{\mathrm{G}}(30)
  = \frac{1}{S}\sum_{s=1}^{S}
    \sqrt{\frac{1}{n_s}\sum_{i=1}^{n_s} e_{si6}^{2}}.
  \label{eq:glucofm-ohio-30}
\end{equation}
CGM-LSM instead scores the whole 0--30-minute trajectory, takes the square
root for every origin separately and averages the origin-level scores, so
that participants with more origins weigh more:
\begin{equation}
  R_{\mathrm{L}}(0{:}30)
  = \frac{1}{N}\sum_{s,i}
    \operatorname{round}_{2\mathrm{dp}}
    \left(\sqrt{\frac{1}{6}\sum_{j=1}^{6}e_{sij}^{2}}\right).
  \label{eq:cgmlsm-ohio-30}
\end{equation}
The six steps include the 5- to 25-minute points, which are easier to
predict than the 30-minute endpoint, and a mean of per-origin roots is
smaller than the root of a pooled mean; the 0--60 and 0--120-minute scores
are the same calculation over the first 12 and all 24 of the 24 predicted
steps.  On the same 9,214 CGM-LSM origins and the same persistence
forecasts, the CGM-LSM calculation gives 11.20~mg/dL and the GlucoFM-Bench
endpoint calculation 23.25~mg/dL, so the metric alone produces a gap of the
observed size.

The populations differ as well.  CGM-LSM requires a complete 289-sample
history and a complete 24-sample future and screens both windows for
repeated and extreme values, so its origin selection reads the target
(9,287 windows meet the contiguity requirement and the screen removes 73,
leaving 9,214); it also floors timestamps to five-minute bins, averages
duplicate readings, clips glucose to $[40,400]$~mg/dL and casts the result
to an integer.  GlucoFM-Bench rounds timestamps to the nearest five-minute
point, keeps the first duplicate, interpolates gaps shorter than one hour,
uses a 144-sample history and builds an origin set per horizon, which gives
23,595 eligible 30-minute origins.  The parenthesized uncertainties differ
too: GlucoFM-Bench reports the standard deviation across participant RMSEs
and CGM-LSM a Student-$t$ 95\% confidence half-width across origin-level
scores.  Each value is therefore comparable only within its own protocol,
which is how Section~\ref{sec:results-benchmarks} uses
Table~\ref{tab:appendix-cgmlsm} and the GlucoFM-Bench OhioT1DM column.

\clearpage

\section{The Same Comparisons under MAE}
\label{sec:additional-mae}

Every result in the paper is reported as RMSE, the metric all three
benchmarks use, and because RMSE weights large errors heavily we repeat the
comparisons under mean absolute error.  On CGMacros the ordering is
unchanged at 30, 60 and 90 minutes: \agentname{} with the multimodal context has
the lowest MAE.  At 120 minutes the multimodal foundation-model baseline
is slightly ahead under MAE, which matches how close those two cells
already are under RMSE.  On OhioT1DM under the CGM-LSM protocol the agent is within
0.35~mg/dL of the foundation models under MAE in every window, as it is
under RMSE, and on GlucoFM-Bench it has the lowest MAE on four of the
nine datasets.

\begin{table}[htbp]
  \caption{CGMacros HOLDOUT MAE sensitivity analysis on the full
  1,903-origin set. Cells report participant-level MAE mean (sample standard
  deviation), in mg/dL; standard deviations use $\mathrm{ddof}=1$. All
  12-hour rows use 144 five-minute history steps.}
  \label{tab:appendix-cgmacros-holdout-mae}
  \centering
  \scriptsize
  \setlength{\tabcolsep}{1.8pt}
  \resizebox{\textwidth}{!}{%
  \begin{tabular}{llrrrr}
    \toprule
    Model & Input setting & 30 min & 60 min & 90 min & 120 min \\
    \midrule
    Persistence & Last observed CGM & 13.512 (3.827) & 19.710 (7.136) & 23.878 (9.796) & 26.614 (11.706) \\
    \addlinespace
    \multirow{2}{*}{TimesFM-2.5}
      & CGM only (12 h) & \best{11.729} (3.125) & \best{17.133} (6.332) & \best{20.196} (8.623) & \best{21.978} (10.029) \\
      & CGM only (full history) & \runnerup{11.386} (3.157) & \runnerup{16.156} (6.031) & 18.604 (8.131) & 19.712 (9.488) \\
    \addlinespace
    \multirow{2}{*}{Moirai-2.0}
      & CGM only (12 h) & 12.322 (3.726) & 18.381 (6.919) & 21.818 (9.790) & 23.756 (11.455) \\
      & CGM only (full history) & 11.521 (3.301) & 16.316 (5.910) & 18.687 (7.854) & 20.042 (9.123) \\
    \addlinespace
    \multirow{2}{*}{Timer$^{\dagger}$}
      & CGM only (12 h) & 13.618 (4.427) & 18.974 (7.863) & 21.930 (10.231) & \runnerup{22.990} (11.572) \\
      & CGM only (full history) & 12.361 (3.775) & 17.141 (6.585) & 19.470 (8.412) & 20.848 (9.745) \\
    \addlinespace
    \multirow{3}{*}{Chronos-2}
      & CGM only (12 h) & \runnerup{12.135} (3.160) & \runnerup{18.112} (6.474) & \runnerup{21.250} (8.705) & 23.042 (10.406) \\
      & CGM only (full history) & \best{11.381} (3.237) & \best{15.998} (5.867) & \best{18.054} (7.613) & \best{19.002} (8.586) \\
      & Multimodal (full history) & \runnerup{11.416} (3.444) & \runnerup{15.834} (5.943) & \runnerup{17.881} (7.738) & \best{18.490} (8.453) \\
    \midrule
    \multirow{3}{*}{\agentname{}}
      & CGM only (12 h) & 12.414 (3.157) & 18.182 (5.968) & 21.360 (8.181) & 23.279 (9.833) \\
      & CGM only (full history) & 11.613 (3.182) & 16.363 (5.454) & \runnerup{18.426} (6.990) & \runnerup{19.333} (7.971) \\
      & Multimodal (full history) & \best{10.905} (2.559) & \best{15.153} (4.988) & \best{17.744} (6.681) & \runnerup{18.653} (7.548) \\
    \bottomrule
  \end{tabular}}
  \vspace{2pt}

  {\scriptsize \textbf{Bold} marks the lowest MAE and
  \underline{underlining} the second-lowest within each exactly comparable
  input setting; only means are styled. $^{\dagger}$Timer
  preserves its upstream 96-step patch flooring (so 12 hours becomes 8
  hours) and uses causal-mean imputation for missing historical CGM.}
\end{table}

\begin{table}[htbp]
  \caption{GlucoFM-Bench MAE sensitivity analysis on the largest completed
  exact local subset ($K=32$ origins per eligible participant): 12-hour
  context, 30-minute horizon, and participant-level MAE in mg/dL. Values are
  mean (sample standard deviation).}
  \label{tab:appendix-glucofm-30min-mae}
  \centering
  \scriptsize
  \setlength{\tabcolsep}{2.4pt}
  \resizebox{\textwidth}{!}{%
  \begin{tabular}{lrrrrrr}
    \toprule
    Dataset & Persistence & Chronos-2 & Moirai-2.0 & Timer & TimesFM-2.5 & \agentname{} \\
    \midrule
    Hall2018
      & 10.11 (5.94) & 9.54 (4.83) & \runnerup{9.39} (5.03) & 10.61 (5.62) & \best{9.27} (4.60) & 9.68 (4.30) \\
    BIG IDEAS Lab
      & 10.20 (2.75) & \runnerup{9.56} (2.49) & 9.59 (2.50) & 10.72 (3.08) & \best{9.17} (2.36) & 9.79 (2.61) \\
    ShanghaiT1DM
      & 13.46 (5.91) & 9.52 (3.69) & 9.03 (3.18) & 15.40 (7.00) & \runnerup{8.89} (3.39) & \best{8.12} (2.68) \\
    ShanghaiT2DM
      & 10.80 (3.97) & 8.38 (2.85) & 8.34 (2.72) & 11.05 (4.11) & \best{7.93} (2.69) & \runnerup{8.00} (2.72) \\
    HUPA-UCM
      & 15.62 (6.08) & 12.42 (5.34) & \runnerup{11.97} (4.90) & 18.02 (5.98) & 12.06 (5.37) & \best{10.97} (4.32) \\
    T1DM-UOM
      & 18.91 (4.77) & 16.43 (3.80) & 16.61 (3.86) & 19.16 (5.18) & \runnerup{16.28} (3.89) & \best{15.78} (3.82) \\
    Bris-T1D Open
      & 13.76 (7.95) & 11.55 (6.59) & 11.49 (6.49) & 14.43 (8.03) & \runnerup{11.28} (6.58) & \best{11.15} (6.63) \\
    AZT1D
      & 16.78 (4.53) & 14.96 (4.16) & 14.80 (4.21) & 17.69 (4.25) & \best{14.33} (3.91) & \runnerup{14.36} (4.18) \\
    OhioT1DM
      & 17.17 (3.25) & \best{14.24} (2.62) & \runnerup{14.27} (2.52) & 18.85 (3.50) & 14.61 (2.22) & 14.62 (3.82) \\
    \bottomrule
  \end{tabular}}
  \vspace{2pt}

  {\scriptsize \textbf{Bold} is best and \underline{underlining} is
  second best on the exact matched subset; only means are styled.}
\end{table}

\begin{table}[htbp]
  \caption{OhioT1DM MAE sensitivity analysis under the CGM-LSM-matched
  protocol on the evaluated subset ($n=1536$). Cells report mean
  per-instance cumulative-window MAE (Student-$t$ 95\% confidence
  half-width), in mg/dL.}
  \label{tab:appendix-cgmlsm-mae}
  \centering
  \small
  \setlength{\tabcolsep}{6pt}
  \begin{tabular}{lccc}
    \toprule
    Method & 0--30 min & 0--60 min & 0--120 min \\
    \midrule
    Chronos-2 (zero-shot) & \runnerup{7.857} (0.352) & \runnerup{13.507} (0.585) & \runnerup{21.804} (0.860) \\
    TimesFM-2.5 (zero-shot) & \best{7.688} (0.351) & \best{13.322} (0.578) & \best{21.554} (0.845) \\
    Persistence & 9.783 (0.443) & 16.148 (0.685) & 25.374 (0.985) \\
    \midrule
    \agentname{} & 7.953 (0.362) & 13.658 (0.596) & 21.893 (0.875) \\
    \bottomrule
  \end{tabular}
  \vspace{2pt}

  {\footnotesize Parenthesized values are confidence half-widths rather than
  standard deviations. \textbf{Bold} is best and \underline{underlining}
  is second best among the exactly matched rows; only means are styled. The
  CGM-LSM publication row is omitted because the publication does not report
  MAE.}
\end{table}

\begin{table}[htbp]
  \caption{Multimodal LLM MAE ablation on the full 1,903-origin CGMacros
  HOLDOUT set. Cells report participant-level MAE mean (sample standard
  deviation), in mg/dL. Mean-4 is computed per participant across the four
  horizons before aggregation.}
  \label{tab:appendix-llm-multimodal-mae}
  \centering
  \scriptsize
  \setlength{\tabcolsep}{3.2pt}
  \resizebox{\textwidth}{!}{%
  \begin{tabular}{llrrrrr}
    \toprule
    Model & Effort & 30 min & 60 min & 90 min & 120 min & Mean-4 \\
    \midrule
    Claude Opus 5 (ours) & high & \best{10.905} (2.559) & \best{15.153} (4.988) & \best{17.744} (6.681) & \best{18.653} (7.548) & \best{15.614} (5.345) \\
    Claude Opus 5 & medium & 11.570 (2.621) & 16.310 (5.195) & 18.869 (6.965) & 20.344 (8.195) & 16.773 (5.628) \\
    Claude Sonnet 5 & high & 12.510 (2.950) & 18.532 (5.551) & 21.774 (7.781) & 24.288 (9.443) & 19.276 (6.310) \\
    Gemini 3.1 Pro & high & 12.596 (3.424) & 17.388 (6.251) & 20.289 (7.312) & 21.317 (8.963) & 17.898 (6.401) \\
    GPT-5.6-sol & high & \runnerup{11.495} (2.983) & \runnerup{15.978} (5.366) & \runnerup{18.519} (6.489) & \runnerup{19.210} (7.473) & \runnerup{16.300} (5.484) \\
    \bottomrule
  \end{tabular}}
  \vspace{2pt}

  {\scriptsize \textbf{Bold} is best and \underline{underlining} is
  second best on the full matched set; only means are styled.}
\end{table}

\begin{table}[htbp]
  \caption{Claude-family MAE ablation across matched input settings on the
  full 1,903-origin CGMacros HOLDOUT set. Cells report participant-level MAE
  mean (sample standard deviation), in mg/dL. Mean-4 is computed per
  participant across the four horizons before aggregation.}
  \label{tab:appendix-llm-claude-mae}
  \centering
  \scriptsize
  \setlength{\tabcolsep}{2.4pt}
  \resizebox{\textwidth}{!}{%
  \begin{tabular}{lllrrrrr}
    \toprule
    Input setting & Model & Effort & 30 min & 60 min & 90 min & 120 min & Mean-4 \\
    \midrule
    \multirow{3}{*}{Multimodal} & Claude Opus 5 (ours) & high & \best{10.905} (2.559) & \best{15.153} (4.988) & \best{17.744} (6.681) & \best{18.653} (7.548) & \best{15.614} (5.345) \\
     & Claude Opus 5 & medium & \runnerup{11.570} (2.621) & \runnerup{16.310} (5.195) & \runnerup{18.869} (6.965) & \runnerup{20.344} (8.195) & \runnerup{16.773} (5.628) \\
     & Claude Sonnet 5 & high & 12.510 (2.950) & 18.532 (5.551) & 21.774 (7.781) & 24.288 (9.443) & 19.276 (6.310) \\
    \addlinespace
    \multirow{3}{*}{CGM only (full history)} & Claude Opus 5 (ours) & high & \best{11.613} (3.182) & \best{16.363} (5.454) & \best{18.426} (6.990) & \best{19.333} (7.971) & \best{16.434} (5.804) \\
     & Claude Opus 5 & medium & \runnerup{12.447} (3.071) & \runnerup{17.703} (5.910) & \runnerup{19.985} (8.265) & \runnerup{21.376} (9.515) & \runnerup{17.878} (6.589) \\
     & Claude Sonnet 5 & high & 12.648 (3.254) & 18.250 (5.727) & 21.958 (8.109) & 24.359 (9.704) & 19.303 (6.603) \\
    \addlinespace
    \multirow{3}{*}{CGM only (12 h)} & Claude Opus 5 (ours) & high & \best{12.414} (3.157) & \best{18.182} (5.968) & \best{21.360} (8.181) & \best{23.279} (9.833) & \best{18.809} (6.667) \\
     & Claude Opus 5 & medium & 13.115 (3.130) & \runnerup{19.406} (5.883) & \runnerup{22.234} (8.436) & \runnerup{23.880} (10.225) & \runnerup{19.659} (6.776) \\
     & Claude Sonnet 5 & high & \runnerup{13.035} (3.556) & 19.599 (6.448) & 23.530 (9.086) & 26.820 (10.955) & 20.746 (7.366) \\
    \bottomrule
  \end{tabular}}
  \vspace{2pt}

  {\scriptsize \textbf{Bold} is best and \underline{underlining} is
  second best within each input setting on the full matched set; only means
  are styled.}
\end{table}

\begin{table}[htbp]
  \caption{MAE comparison with the hand-designed tool interface on the full 1,903
  exactly matched CGMacros HOLDOUT origins. Cells report participant-level
  MAE mean (sample standard deviation), in mg/dL. Mean-4 is computed per
  participant across the four horizons before aggregation.}
  \label{tab:appendix-hand-design-multimodal-mae}
  \centering
  \scriptsize
  \setlength{\tabcolsep}{3.0pt}
  \resizebox{\textwidth}{!}{%
  \begin{tabular}{lrrrrr}
    \toprule
    Method & 30 min & 60 min & 90 min & 120 min & Mean-4 \\
    \midrule
    Hand-designed faithful port (high effort) & 11.877 (2.528) & 16.520 (5.103) & 18.709 (7.314) & 19.758 (8.622) & 16.716 (5.770) \\
    Hand-designed faithful port (medium effort) & 12.317 (2.889) & 17.399 (5.566) & 19.818 (7.944) & 20.970 (9.475) & 17.626 (6.303) \\
    Chronos-2 (fuller multimodal) & \runnerup{11.416} (3.444) & \runnerup{15.834} (5.943) & \runnerup{17.881} (7.738) & \best{18.490} (8.453) & \runnerup{15.905} (6.307) \\
    \agentname{} (agent-designed multimodal) & \best{10.905} (2.559) & \best{15.153} (4.988) & \best{17.744} (6.681) & \runnerup{18.653} (7.548) & \best{15.614} (5.345) \\
    \bottomrule
  \end{tabular}}
  \vspace{2pt}

  {\scriptsize \textbf{Bold} marks the lowest MAE and
  \underline{underlining} the second-lowest in each column; only means are
  styled.}
\end{table}

\end{document}